\documentclass[10pt,twocolumn,letterpaper]{article}

\usepackage[pagenumbers]{iccv} 

\usepackage{graphicx}
\usepackage{colortbl}
\usepackage{multirow}
\usepackage{booktabs}
\usepackage{xcolor}
\usepackage{wrapfig,lipsum}
\usepackage{subcaption}
\usepackage{xspace}

\usepackage{amsmath,amsfonts,bm}

\def\eqref#1{equation~\ref{#1}}

\def\1{\bm{1}}

\DeclareMathAlphabet{\mathsfit}{\encodingdefault}{\sfdefault}{m}{sl}
\SetMathAlphabet{\mathsfit}{bold}{\encodingdefault}{\sfdefault}{bx}{n}

\def\gD{{\mathcal{D}}}

\def\gN{{\mathcal{N}}}

\newcommand{\ours}{LoFT\xspace}
\newcommand{\ourslong}{\textbf{Lo}RA-\textbf{F}used \textbf{T}raining-data Generation with Few-shot Guidance\xspace}

\newcommand{\myparagraph}[1]{\noindent{\bf{#1}}}

\definecolor{iccvblue}{rgb}{0.21,0.49,0.74}
\usepackage[pagebackref,breaklinks,colorlinks,allcolors=iccvblue]{hyperref}

\def\paperID{8946} 
\def\confName{ICCV}
\def\confYear{2025}

\title{LoFT: LoRA-Fused Training Dataset Generation with Few-shot Guidance}

\author{
Jae Myung Kim$^{1,2,4}$
\;\;\;\; Stephan Alaniz$^{5}$
\;\;\;\; Cordelia Schmid$^{6}$
\;\;\;\; Zeynep Akata$^{2,3,4}$ 
\\
\\
\small{$^{1}$University of Tübingen}
\;\;\;\; \small{$^{2}$Helmholtz Munich}
\;\;\;\; \small{$^{3}$Technical University of Munich}
\;\;\;\; \small{$^{4}$Munich Center for Machine Learning}  \\
\small{$^{5}$LTCI, Télécom Paris, Institut Polytechnique de Paris, France}
\;\;\;\; \small{$^{6}$Inria, Ecole normale sup\'erieure, CNRS, PSL Research University}\\
}

\begin{document}
\maketitle

\begin{abstract}
Despite recent advances in text-to-image generation, using synthetically generated data seldom brings a significant boost in performance for supervised learning. Oftentimes, synthetic datasets do not faithfully recreate the data distribution of real data, i.e., they lack the fidelity or diversity needed for effective downstream model training. While previous work has employed few-shot guidance to address this issue, existing methods still fail to capture and generate features unique to specific real images. In this paper, we introduce a novel dataset generation framework named \ours, \ourslong. Our method fine-tunes LoRA weights on individual real images and fuses them at inference time, producing synthetic images that combine the features of real images for improved diversity and fidelity of generated data. We evaluate the synthetic data produced by \ours on 10 datasets, using 8 to 64 real images per class as guidance and scaling up to 1000 images per class. Our experiments show that training on \ours-generated data consistently outperforms other synthetic dataset methods, significantly increasing accuracy as the dataset size increases. Additionally, our analysis demonstrates that \ours generates datasets with high fidelity and sufficient diversity, which contribute to the performance improvement. The code is available at \href{https://github.com/ExplainableML/LoFT}{https://github.com/ExplainableML/LoFT}.
\end{abstract}

\begin{figure}[t]
    \centering
    \includegraphics[width=\linewidth]{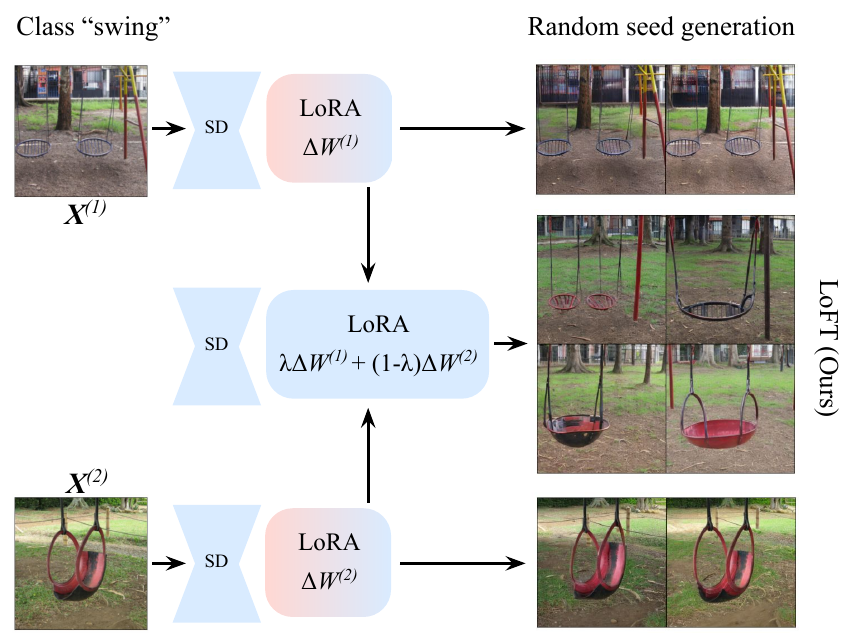}
    \caption{
    LoFT: Given a few real images per class, we first adapt a diffusion model to each image using LoRA. Next, two LoRA weights corresponding to images of the same class are randomly selected and fused to generate new images. The generated synthetic images above show diverse colors and compositions while maintaining the swing object.
    } 
    \label{fig:teaser}
\end{figure}

\section{Introduction}

Synthetic data offers a cost-effective alternative to the labor-intensive process of real data collection. One promising downstream application of diffusion-based text-to-image generative models~\cite{ddpm,stable_diffusion, sdxl,dalle2,dalle3,imagen} is to augment real datasets with synthetic images~\cite{diversity2,diversity3} or training models on entirely synthetic data~\cite{issynth,fakeit,diversity1}. 
While these methods show potential, models trained solely on synthetic data often underperform compared to those trained on real data~\cite{scalelaw}. This is largely due to distributional misalignment between synthetic and real data, as well as a lack of fine-grained detail in the generated images~\cite{datadream, scalelaw}. 

To tackle distribution shift issue, recent works suggests to guide the dataset generation with a few real data samples~\cite{issynth,disef,datadream}. In this few-shot setting, we assume to have access to a few real images for every class for an image classification task. Yet, the issue of misalignment remains a challenge for certain classes or downstream datasets.
For instance, \citet{disef} use partially noised real image as conditional input to the diffusion model and generate synthetic data using prompts from a captioning model. However, these images often deviate from the real data distribution, making them less relevant to the task at hand. \citet{datadream} propose DataDream, which fine-tunes the diffusion model with few-shot data to learn the data distribution, but we find that it struggles to generate in-distribution images for all classes consistently. This is because DataDream finetunes on all available images from the same class, which makes it challenging to retain high-fidelity details of individual images (e.g., less frequently visible parts of a class, such as the back of a car), focusing instead on commonly shared features. 

We introduce \ours, \ourslong to generate high-fidelity, in-distribution synthetic images using few-shot real images. 
Instead of fine-tuning a diffusion model on all image of a class jointly, we train separate sets of Low-Rank Adaptation (LoRA) parameters on individual images, i.e., the diffusion model learns to overfit to a single image, generating it exclusively.
At inference time, we then fuse together the LoRA weights of any two real images from the same class, to generate synthetic images that share the characteristics of both images.
As shown in Figure~\ref{fig:teaser}, given two images of swings, the individual LoRA weights lead to generations similar to the real image (top and bottom), while the fused LoRA weights create images inheriting features from both source images
while still maintaining the identity of a swing.
There are two advantages of our LoFT method. First, learning separate LoRA weights for each individual image eases the diffusion model adaptation as the finetuning can retain on every detail of the real image. This instance-level adaptation ensures better alignment between the distribution of the few-shot real images and the synthetic images generated by the LoRA-tuned diffusion model, resulting in high fidelity. Second, by fusing the LoRA weights from different images of the same class, we maintain the diversity of the generated synthetic images.

Our key contributions are:
(1) introducing \ours, a few-shot guided synthetic dataset generation method that generates high-fidelity, in-distribution synthetic datasets by training LoRA adapters per image and fusing them when generating synthetic images;
(2) providing a comprehensive comparison of four synthetic dataset generation methods on ten downstream datasets, demonstrating superior performance in fine-tuning CLIP when trained on data from \ours; and
(3) analyzing synthetic data generation methods based on fidelity and diversity, showing that \ours achieves high fidelity with sufficient diversity, leading to improved performance when using its synthetic dataset.

\section{Related Work}

Diffusion-based text-to-image (T2I) models have enabled the creation of highly realistic synthetic images~\cite{stable_diffusion,sdxl,dalle2,dalle3,imagen,nichol2021glide}. These models operate by gradually denoising Gaussian noise,
conditioned on textual prompts. 
Promising downstream applications of T2I generative models include generating training data for 
classification~\cite{issynth,fakeit,datadream,diversity1,diversity2,diversity3,thinair,lei2023image,azizi2023synthetic,zheng2023toward,zhang2023prompt,yuan2024realfake,fu2024dreamda,yeo2024controlled,tan2024semantic}, handling long-tail distributions~\cite{shin2023fill,ye2023synaug,hemmat2023feedback}, data distribution shifts~\cite{distribution_shift,du2023dream,liu2025does}, semi-supervised learning~\cite{you2023diffusion}, representation learning~\cite{stablerep}, object detection~\cite{object_detection_synth}, vision-language pre-training~\cite{synclr,synthclip,sharifzadeh2024synth}, and image generation~\cite{mad,bertrand2024stability}.
For image classification, a lot of work has focused on generating synthetic images zero-shot, which typically involves generating data from text prompts that include the class names from the downstream task \cite{issynth,fakeit,diversity1,diversity2,zhang2023prompt}. However, this approach often leads to generated images that lack a faithful representation of the target object, resulting in a mismatch between synthetic and real images, which hinders performance gains \cite{datadream,synclr}. To mitigate these issues, there has been growing interest in few-shot learning, where limited real data is used alongside synthetic data. Techniques such as initializing the generation from a partially noised real image~\cite{issynth,disef} or fine-tuning the diffusion model with few-shot data~\cite{datadream} have been employed to align the synthetic data more closely with real-world distributions. 
Our method differs from other few-shot guided methods by using LoRA fusion to combine the features of multiple real images.

Controllable text-to-image diffusion models have enabled personalization in image generation~\cite{textual_inv,dreambooth,hyperdreambooth,antidreambooth,qiu2023controlling,instantbooth}.
Fine-tuning a diffusion model with LoRA~\cite{lora} and fusing them in the image generation phase has been demonstrated to be an effective technique for image morphing~\cite{zhang2024diffmorpher} and model customization~\cite{dravid2024interpreting}, where LoRA weights are fused to achieve customized outputs. In contrast, our method leverages LoRA fusion for synthetic dataset generation and demonstrates its effectiveness in training classification models. While \citet{thinair} proposed fusing learned tokens from Textual Inversion~\cite{textual_inv} for synthetic dataset generation, we show
that our \ours outperforms these previous fusion methods.

To understand and improve the impact of synthetic training data, \citet{scalelaw} measure the fidelity and diversity of synthetic datasets. Fidelity can be improved through methods like CLIP filtering~\cite{diversity3,issynth,object_detection_synth} and incorporating additional class information~\cite{fakeit}, while diversity is affected by the guidance scale~\cite{fakeit,scalelaw}, adding attributes to prompts~\cite{diversity3,diversity2,fakeit}, or using large language models to generate more varied prompts~\cite{synthclip,issynth}.
Additionally, \citet{scalelaw} have explored scaling laws in synthetic training data, demonstrating that synthetic datasets do not exhibit the same scaling benefits as real data in supervised tasks. In this work, we examine few-shot guided dataset generation methods in terms of both fidelity and diversity, and further investigate how scaling up synthetic datasets to sizes of up to one million affects the performance of these methods.

\begin{figure*}
    \centering
    \includegraphics[width=\linewidth]{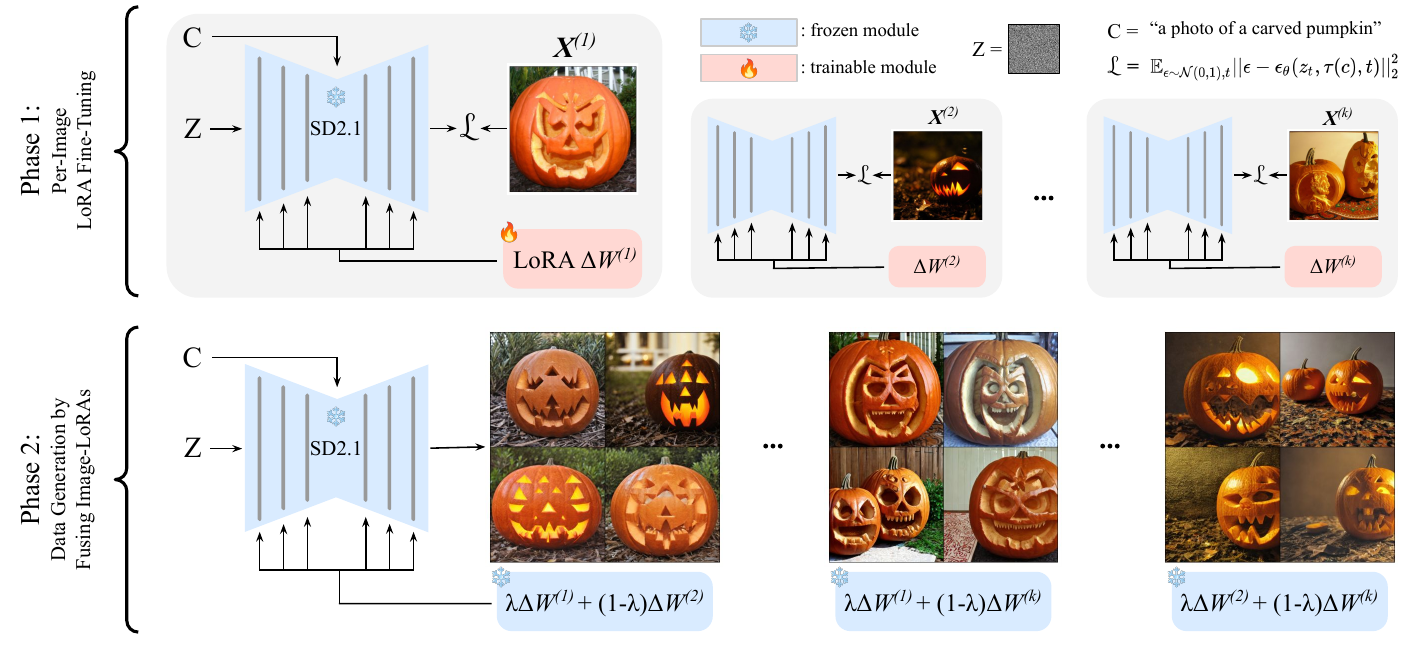}
    \caption{\ours pipeline. In the first phase, given a few real images per class, we adapt a diffusion model to each image using LoRA. In the second phase, two LoRA weights corresponding to images of the same class are randomly selected and fused to generate new synthetic images. These generated images are then compiled to form a dataset for training the classification model.
    }
    \label{fig:pipeline}
\end{figure*}

\section{LoRA-Fused Training Dataset Generation}

In this section, we begin by describing baseline methods for synthetic dataset generation in the zero-shot and few-shot scenarios (\S\ref{method:zero-few-shot}). We then introduce our proposed method, \ours, in \S\ref{method:ours}.

\subsection{Synthetic dataset generation}
\label{method:zero-few-shot}

\myparagraph{Stable diffusion: text-to-image generation.} 
The Stable Diffusion~\cite{stable_diffusion} model learns a conditional probability distribution $p(x|c)$ given a data point $(x,c) \in \gD$ where $x$ is an image and $c$ is its caption. The model learns a reverse process of gradually denoising Gaussian noise in the latent space. Concretely, the diffusion and reverse processes work in a latent space, which is defined through a pre-trained image encoder $f$ that encodes the image $x$ to a latent $z$, i.e. $z\!=\!f(x)$, and the corresponding decoder $g$ where $x\!=\!g(z)$. Given a time step $t \in \{0, \cdots\!, T\}$, $z_t$ denotes noisy latent state after $t$ steps of small Gaussian noise addition from $z_0\!=\!z$ where $z_T$ is Gaussian noise. The latent diffusion models' objective is to minimize the following loss:
\begin{equation}
    \min_{\theta} \,\, \mathbb{E}_{(x,c) \sim \gD, \, \epsilon \sim \gN(0,1), \, t} \, \left[\, \left\| \, \epsilon - \epsilon_{\theta} (z_t, \tau(c), t) \, \right\|_2^2 \,\right] \, ,
\end{equation}
where $\tau(\cdot)$ is a text encoder. Intuitively, the loss enables the model
to learn to denoise the latent $z_t$. During inference, we start with noise $z_T$ and iteratively denoise it through $T$ steps of the latent diffusion model, obtaining $z_0$. This latent is decoded by the pre-trained decoder $g$ to generate a final image $x'=g(z_0)$.

\paragraph{Zero-shot image generation.} 
New images can be generated by conditioning the model on a template prompt~\cite{issynth,fakeit}, such as \texttt{"a photo of a \{$l$\}"}, where $l$ represents a class name. As a result, the synthetic dataset $$\gD^{\text{synth}}=\{(x_i, y_i)\}_{i=1}^{sL}$$ contains $s$ generated images for each of the $L$ classes where every image $x_i$ is automatically annotated by the class label $y_i \in \{1,2,\cdots\!, L\}$ derived from its textual prompt. To improve diversity in these generated images, lowering the guidance scale has been shown to be effective, as it encourages more output variety, therefore improving classification performance \cite{fakeit,scalelaw}. We refer to this method as \textbf{ClassPrompt}.

\paragraph{Few-shot guided dataset generation.}
While zero-shot text-to-image methods can generate a large amount of distinct images, they often struggle to produce the classification object of interest or capture fine-grained details of a class~\cite{datadream}. To address this, few-shot guided approaches have been developed, where we assume access to a few real images for each class. In the $k$-shot setting, we denote $$\gD^{\text{fs}}=\{(x_i, y_i)\}_{i=1}^{kC}$$ as the few-shot dataset, where $x_i$ is an image, $y_i$ is the label of the image, $k$ is the number of available real images per class, and $C$ is the number of classes. Few images per class can already provide rich visual information to better inform the data generation process beyond textual labels alone, while not requiring an extensive effort to collect.
For instance, 
DataDream~\cite{datadream} fine-tunes LoRA weights applied on the diffusion model using few-shot real images. 
In our experiments, we use DataDream with LoRA weights trained for each class as a representative of the few-shot guided image generation approach based on fine-tuning, and refer to this as \textbf{DataDream}.

While lowering the guidance scale increases variety in zero-shot image generation, using a template text prompt still limits the generation of a diverse dataset. To further increase diversity, \citet{diversity1} have leveraged large language models (LLMs) to enrich prompts with additional context or attributes related to the class name. 
Additionally, \citet{diversity3} and \citet{scalelaw} leverage real images by applying a captioning model to create detailed captions from these images, which are then used as prompts for generation.
In our experiments, we include a baseline for few-shot guided data generation through captioning. Specifically, we caption the few-shot images for each class with PaliGemma~\cite{paligemma},
a multimodal large language model. We generate one caption per real image, i.e., $k$ captions per class in the $k$-shot setting. We then use these captions as prompts for synthetic image generation. We refer to this method as \textbf{CaptionPrompt}.

\subsection{\ours method}
\label{method:ours}

While DataDream has shown promising results for few-shot guided dataset generation, we find that it
struggles to generate in-distribution images for some classes,
limiting its impact on classification performance.
This issue arises when there is high diversity in the few-shot images such that the fine-tuned diffusion models do not faithfully represent fine-grained details that may occur only in one of the images, leading to underfitting.

To overcome these challenges, we propose \ours, \ourslong for generating better in-distribution synthetic images. As shown in Figure~\ref{fig:pipeline}, \ours fine-tunes the pre-trained diffusion model with one set of LoRA weights for \textit{every real image} $x_i$ from the few-shot dataset $\gD^{\text{fs}}$ independently. Specifically, for every attention layer of the diffusion model U-net, we add LoRA~\cite{lora} parameters to the linear weight matrices
\begin{equation}
    h_{\text{out}} = W h_{\text{in}} +  \Delta W^{(i)} h_{\text{in}}
\end{equation}
where $h$ is the activation of a linear layer, $W \in \mathbb{R}^{d_1 \times d_2}$ is the original weight matrix, and $\Delta W^{(i)}$ is the low rank adaptation matrix which is optimized. The parameterization $$\Delta W^{(i)} = B^{(i)} A^{(i)}$$ with $B \in \mathbb{R}^{d_1 \times r}$ and $A \in \mathbb{R}^{r \times d_2}$ allows for efficient fine-tuning because the low rank $r \ll min(d_1, d_2)$ reduces the number of tunable parameters significantly. The parameter efficiency and modularity of LoRA lead to a low storage costs and flexible inference-time manipulation (fusion).

\paragraph{Fine-tuning LoRA weights on a single image.} As presented in the grey box in Figure~\ref{fig:pipeline}, we fine-tune a separate set of LoRA weights $\Delta W^{(i)}$ for each data point $(x_i, y_i)$ with the diffusion model objective while keeping the original parameters fixed:
\begin{equation}
    \min_{\Delta W^{(i)}} \,\, \mathbb{E}_{\epsilon \sim \gN(0,1), \, t} \, \left[\, || \, \epsilon - \epsilon_{\theta\!, \Delta W^{(i)}} (z_t, \tau(C(y_i)), t) \, ||_2^2 \,\right] \, , 
    \label{eq:datadream_loss}
\end{equation}
where $z_t$ corresponds to the noised $x_i$ at step $t$ and $C(y_i)$ is the template prompt \texttt{"a photo of a \{$l_i$\}"} with $l_i$ being the class name of $y_i$.
By learning LoRA weights for every few-shot image individually, \ours overfits the diffusion model to a single sample, learning to reproduce all of its details even if such features were not in the training data of the original diffusion model. At inference time, every generated image will closely resemble the original real image, increasing fidelity, i.e., replicating fine-grained details.

\paragraph{Fusing LoRA weights.} To increase the generation diversity from single-image LoRAs, we propose to interpolate their weights. We fuse two randomly selected LoRA weights corresponding to real images from the same class with
\begin{equation}
    h_{\text{out}} = W h_{\text{in}} + \lambda \Delta W^{(i)} h_{\text{in}} + (1-\lambda) \Delta W^{(j)} h_{\text{in}}
\end{equation}
where $\lambda \in [0, 1]$ and $\{(i,j) | y_i = y_j\}$.

This fusing strategy combines features from the different instances, effectively interpolating between real images in the weight space of the diffusion model, and improving the diversity of the generated images, as shown in the second phase in Figure~\ref{fig:pipeline}.
While $\lambda = 0$ or $\lambda = 1$ are reproducing the real data, choosing $\lambda = 0.5$ best interpolates images to produce new in-distribution samples of both high fidelity and diversity.

\section{Experiments}

\begin{figure*}[t]
    \centering
    \includegraphics[width=\linewidth]{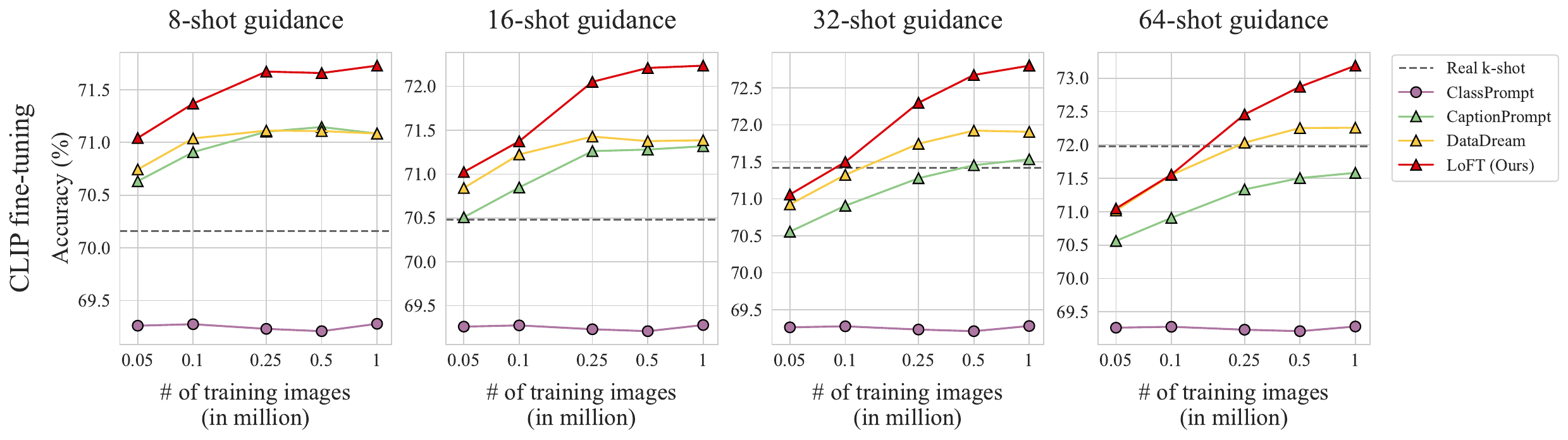}
    \vspace{-15pt}
    \caption{
    Classification accuracy on ImageNet when fine-tuning CLIP on synthetic data generated from different methods at different scales. We report few-shot guidance on 8, 16, 32, and 64 images per class and a baseline of training CLIP only on k-shot real data. \ours consistently outperforms other methods and real k-shot result with small amount of synthetic data. 
    }
    \label{fig:main_result_clip}
\end{figure*}

We use Stable Diffusion~\cite{stable_diffusion} version 2.1 as the generative model for all synthetic dataset generation methods. For the ClassPrompt approach, we utilize the template prompt, \texttt{"a photo of a \{$l$\}"}.
In the CaptionPrompt method, the prompt \texttt{"Caption the image:"}\footnote{Sourced from the \href{https://github.com/google-gemini/gemma-cookbook/blob/main/PaliGemma/Image_captioning_using_PaliGemma.ipynb}{original code base}.} 
is used to caption each input image with PaliGemma~\cite{paligemma}. Once a list of captions is generated, each caption is appended to the template prompt, forming prompts such as \texttt{"a photo of a \{$l$\}, \{caption\}."} These prompts are then used as conditional input to the diffusion model to create synthetic images.
For DataDream, We adopt the hyperparameter configuration \citet{datadream} except that we exclude LoRA on text encoders for training since we found this to perform better.
A guidance scale of 2.0 is used for all methods when generating synthetic images.

For our \ours method, we employ AdamW~\cite{adamw} as the optimizer, a learning rate of 1e-3 with a cosine annealing scheduler, and a LoRA rank $r=2$ for all trained LoRA adapters, which proved sufficient for adapting to single images. When generating images, \ours fuses LoRA weights by randomly selecting two LoRA adaptations and fusing them with equal weights $\lambda=0.5$.
Different fusion strategies of LoRA weights are studied in \S\ref{subsec:ablation}.

\begin{table*}[t]
\centering
\setlength{\tabcolsep}{3.5pt}
\resizebox{0.85\textwidth}{!}{
\renewcommand{\arraystretch}{1.3}
\begin{tabular}{l|ccccccccc|c}
\toprule
Method & Cal & DTD & Eur & Air & Pet & Car & SUN & Food & Flo & Avg \\ \hline \hline
CLIP zero-shot
& 93.0 & 44.4 & 47.6 & 24.7 & 89.2 & 65.2 & 62.6 & 86.1 & 71.4 & 64.9 \\
ClassPrompt
& 93.9$_{\pm0.2}$ & 53.7$_{\pm0.5}$ & 46.2$_{\pm0.7}$ & 26.5$_{\pm0.1}$ & 92.5$_{\pm0.1}$ & 74.1$_{\pm0.5}$ & 67.0$_{\pm0.0}$ & 85.1$_{\pm0.0}$ & 71.9$_{\pm0.5}$ & 67.9$_{\pm0.1}$ \\
CaptionPrompt 
& 95.4$_{\pm0.3}$ & 63.9$_{\pm1.0}$ & 46.6$_{\pm1.7}$ & 26.6$_{\pm0.1}$ & 92.9$_{\pm0.1}$ & 74.4$_{\pm0.1}$ & 73.9$_{\pm0.2}$ & 85.4$_{\pm0.0}$ & 72.3$_{\pm0.5}$ & 70.2$_{\pm0.1}$ \\
DataDream 
& 96.0$_{\pm0.4}$ & 64.9$_{\pm0.2}$ & 84.1$_{\pm3.4}$ & 61.4$_{\pm0.9}$ & \bf 93.5$_{\pm0.2}$ & \bf 90.5$_{\pm0.4}$ & 74.4$_{\pm0.2}$ & \bf 86.5$_{\pm0.0}$ & \bf 98.0$_{\pm0.2}$ & 83.2$_{\pm0.4}$ \\
LoFT (Ours) 
& \bf 96.7$_{\pm0.2}$ & \bf 70.5$_{\pm0.1}$ & \bf 86.8$_{\pm2.2}$ & \bf 66.1$_{\pm1.5}$ & 93.2$_{\pm0.1}$ & 89.3$_{\pm0.5}$ & \bf 75.5$_{\pm0.0}$ & 86.0$_{\pm0.0}$ & \bf 98.0$_{\pm0.2}$ & \bf 84.7$_{\pm0.2}$ \\ %
\bottomrule
\end{tabular}
}
\vspace{-5pt}
\caption{Classification accuracy on 9 fine-grained benchmarks when fine-tuning CLIP on synthetic data with 16-shot guidance. 500 synthetic images are generated for each class. Datasets are Cal: Caltech 101, Eur: EuroSAT, Air: FGVC Aircraft, Flo: Flowers 102.
}
\label{tab:finegrained_synth}
\end{table*}

\begin{table*}[t]
\centering
\setlength{\tabcolsep}{3.5pt}
\resizebox{0.85\textwidth}{!}{
\renewcommand{\arraystretch}{1.3}
\begin{tabular}{l|ccccccccc|c}
\toprule
Method & Cal & DTD & Eur & Air & Pet & Car & SUN & Food & Flo & Avg \\ \hline \hline
CLIP zero-shot~\cite{clip}
& 93.0 & 44.4 & 47.6 & 24.7 & 89.2 & 65.2 & 62.6 & 86.1 & 71.4 & 64.9 \\ 
CoOp~\cite{coop}
& 95.5$_{\pm0.1}$ & 67.8$_{\pm2.2}$ & 78.9$_{\pm0.4}$ & 38.7$_{\pm0.7}$ & 93.3$_{\pm0.3}$ & 78.3$_{\pm0.7}$ & 74.0$_{\pm0.2}$ & 86.7$_{\pm0.6}$ & 95.8$_{\pm0.1}$ & 78.8$_{\pm0.6}$ \\
TIP-Adapter~\cite{tipadapter}
& 95.1$_{\pm0.1}$ & 65.4$_{\pm1.2}$ & 77.6$_{\pm1.0}$ & 39.4$_{\pm0.3}$ & 91.8$_{\pm0.3}$ & 75.6$_{\pm0.5}$ & 72.1$_{\pm0.2}$ & 86.5$_{\pm0.1}$ & 94.6$_{\pm0.1}$ & 77.6$_{\pm0.2}$ \\
TIP-Adapter-f~\cite{tipadapter}
& 95.8$_{\pm0.1}$ & 72.2$_{\pm0.3}$ & 89.0$_{\pm0.4}$ & 44.9$_{\pm0.4}$ & 93.0$_{\pm0.2}$ & 83.3$_{\pm0.5}$ & 76.3$_{\pm0.2}$ & \bf 87.3$_{\pm0.0}$ & 96.8$_{\pm0.2}$ & 82.1$_{\pm0.0}$ \\
AMU-Tuning~\cite{amutuning}
& 97.1$_{\pm0.3}$ & 70.0$_{\pm1.0}$ & 90.4$_{\pm0.4}$ & 47.7$_{\pm1.6}$ & 92.8$_{\pm0.1}$ & 78.5$_{\pm0.1}$ & 72.6$_{\pm0.2}$ & 85.7$_{\pm0.2}$ & 95.4$_{\pm0.2}$ & 81.1$_{\pm0.3}$ \\
LoFT (Ours)
& \bf 97.3$_{\pm0.1}$ & \bf 73.8$_{\pm0.5}$ & \bf 93.1$_{\pm0.9}$ & \bf 71.8$_{\pm1.6}$ & \bf 94.3$_{\pm0.4}$ & \bf 90.7$_{\pm0.3}$ & \bf 77.3$_{\pm0.0}$ & 87.2$_{\pm0.1}$ & \bf 99.2$_{\pm0.0}$ & \bf 87.2$_{\pm0.3}$ \\
\bottomrule
\end{tabular}
}
\vspace{-5pt}
\caption{Comparison between the state-of-the-art few-shot learning methods on 9 fine-grained benchmarks. CLIP ViT-B/16 is used as a base model with a 16-shot setting. Baseline methods use real data, and \ours use real data as well as synthetic data for the training set. Datasets are Cal: Caltech 101, Eur: EuroSAT, Air: FGVC Aircraft, Flo: Flowers 102.
}
\label{tab:fewshot_sota_comparison}
\end{table*}

\subsection{Synthetic training data on ImageNet} 

To evaluate different dataset generation methods, we train an image classification model on each synthetic dataset. The target classification task is ImageNet~\cite{imagenet}, which contains 1,000 classes. For each generation method, we produce 50, 100, 250, 500, and 1,000 images per class, corresponding to dataset sizes of 0.05M, 0.1M, 0.25M, 0.5M, and 1M.
We use these datasets to fine-tuning a pre-trained CLIP~\cite{clip} model and evaluate it on the validation set of ImageNet.  
Our goal is to investigate whether synthetic training data can provide additional useful information to improve the performance of a pre-trained model that already has some knowledge of the downstream task. Following the work of \citet{disef} and \citet{datadream}, we use the pre-trained CLIP ViT-B/16 model~\cite{clip} as the base model, and fine-tune a LoRA (rank 16) applied to both the vision encoder and text encoder with the synthetic training data for ImageNet. We also conducted training ResNet50~\cite{resnet} from scratch, which is shown in Appendix~\ref{app:sec:exp_imagenet_rn50}.
For synthetic data generation methods leveraging few-shot real images, we conduct experiments in 8-, 16-, 32- and 64-shot settings to examine performance under different guidance levels.
The CLIP fine-tuning results are shown in Figure~\ref{fig:main_result_clip}.

\myparagraph{ClassPrompt does not scale.}
ClassPrompt dataset generation (purple line) improves over the baseline CLIP performance of 66.6\%, and fluctuates between 69\% and 70\% as the dataset size increases. It indicates that the ClassPrompt method fails to improve as more images are generated.
Consequently, synthetic images generated by the ClassPrompt method provide minimal additional information to the pre-trained CLIP model and showing no ability to scale.

\myparagraph{Few-shot guided methods outperform ClassPrompt.}
Across all dataset sizes and k-shot settings, all few-shot guided methods ({\footnotesize $\triangle$} markers in Figure~\ref{fig:main_result_clip}) outperform ClassPrompt ({\large $\circ$} markers). This is because few-shot guided methods generate higher-quality images with better diversity (for CaptionPrompt) or higher fidelity (for DataDream and LoFT). We provide a detailed analysis in \S\ref{subsec:analysis} and a qualitative examples in \S\ref{subsec:qual}. This indicates that the inclusion of real-image guidance in the synthetic data generation process significantly improves the quality of the synthetic training dataset for training the downstream model.

\myparagraph{Few-shot guided methods outperform real k-shot.}
The dashed line in each plot represents the performance of a model trained solely on k real images per class. We observe that even with a small number of synthetic images, models trained on a few-shot guided synthetic dataset can easily outperform models trained with real k-shot data. For example, the accuracy of the 16-shot real data is 70.48\%. By generating 50 synthetic images per class (resulting in dataset size 0.05M), the accuracy of the model trained on LoFT dataset can reach 71.02\%.

\myparagraph{LoFT effectively scales across different k-shot settings.}
Unlike ClassPrompt, few-shot guided methods show consistent improvement in performance as the dataset size increases, with LoFT achieving the best performance across all k-shot settings. For instance, in the 16-shot setting, LoFT shows a 1.22\% performance improvement, increasing from 71.02\% to 72.24\% as the dataset size grows from 0.05M to 1M training images, while CaptionPrompt shows a 0.81\% improvement (70.51\% $\rightarrow$ 71.32\%). The difference becomes bigger in the 64-shot setting where LoFT shows a 2.14\% improvement (71.05\% $\rightarrow$ 73.19\%) while CaptionPrompt shows a 1.02\% improvement (70.56\% $\rightarrow$ 71.58\%).

\begin{figure*}
    \centering
    \includegraphics[width=\linewidth]{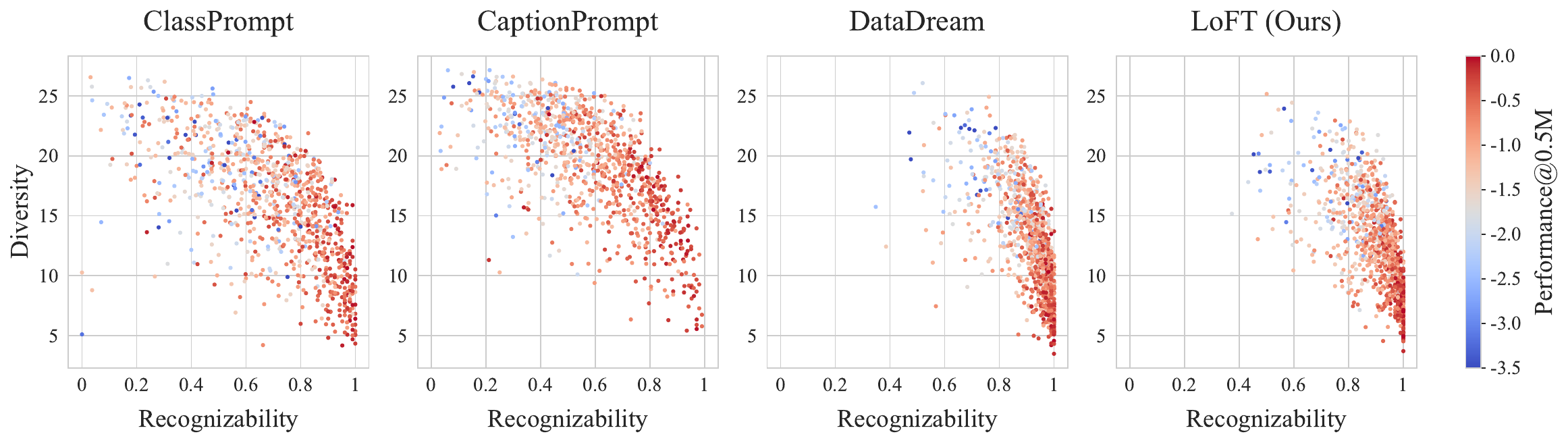}
    \vspace{-15pt}
    \caption{
    Per-class analysis on synthetic datasets generated from different methods.
    The color indicates a log-likelihood of the ImageNet validation dataset when CLIP is fine-tuned on the 0.5M-sized synthetic dataset in the 16-shot setting.
    }
    \label{fig:per_class_analysis}
\end{figure*}

\begin{figure}[t]
    \centering
    \begin{subfigure}{.25\textwidth}
      \centering
      \includegraphics[width=0.95\linewidth]{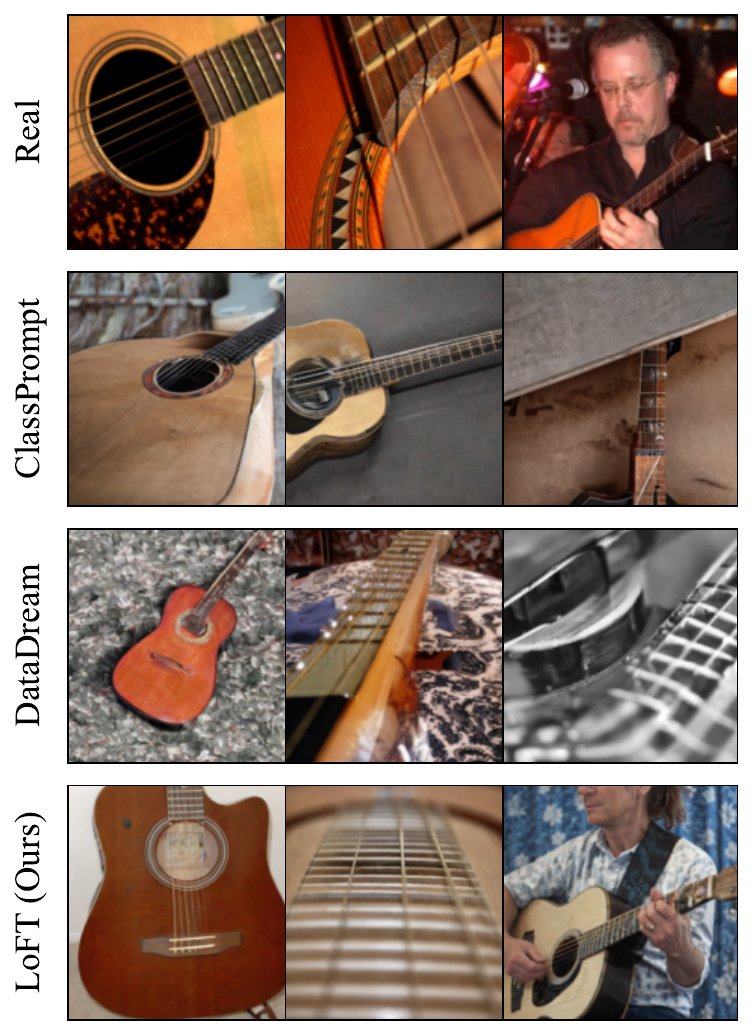}
      \caption{Acoustic guitar}
      \label{fig:qual_sub1}
    \end{subfigure}%
    \begin{subfigure}{.25\textwidth}
      \centering
      \includegraphics[width=0.95\linewidth]{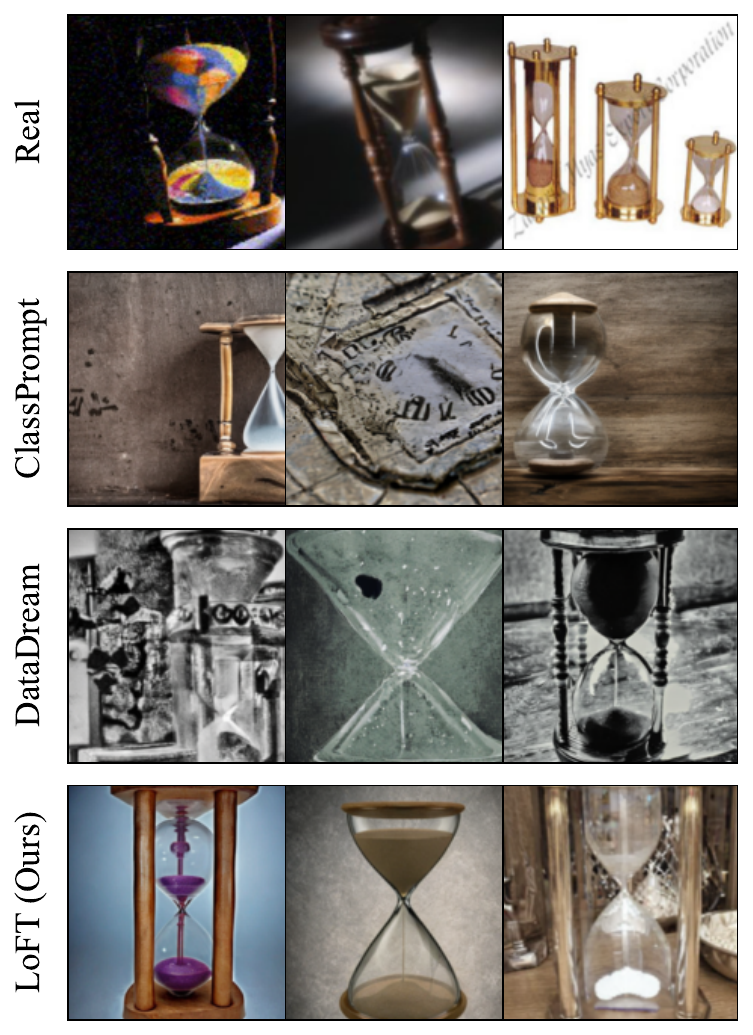}
      \caption{Hourglass}
      \label{fig:qual_sub2}
    \end{subfigure}
    \vspace{-15pt}
    \caption{Qualitative examples for the classes Acoustic guitar and Hourglass from ImageNet. Our LoFT method generates diverse images, such as variations in zoom level, for acoustic guitar, and preserves an object of interest better for hourglass.
    } 
    \label{fig:qualitative_example}
\end{figure}

\subsection{Synthetic training data on fine-grained datasets}

\subsubsection{Comparison of synthetic data generation methods}
To further examine the effectiveness of our \ours method, we evaluate on 9 fine-grained benchmarks: Caltech101~\cite{caltech101}, DTD~\cite{dtd}, EuroSAT~\cite{eurosat}, FGVC Aircraft~\cite{aircraft}, Oxford Pets~\cite{pets}, Stanford Cars~\cite{cars}, SUN397~\cite{sun397}, Food101~\cite{food101}, and Flowers102~\cite{flowers102}.
For each dataset, we generate 500 synthetic images for each class and fine-tune the CLIP model. For few-shot methods, we use 16-shot. 

The results are shown in Table~\ref{tab:finegrained_synth}. We observe that our \ours method outperforms other methods on 6 out of 9 benchmarks, achieving the best average accuracy of 84.7\%. For instance, \ours performs significantly better than DataDream on the DTD dataset (70.5\% vs. 64.9\%) which consists of texture images. This may attribute from LoRA by DataDream having challenges in learning texture patterns with a batch of images, whereas optimizing a single image using LoFT leads to better convergence, thus generating more in-distribution images in the generation phase. Additionally, ClassPrompt and CaptionPrompt underperform on the Aircraft and Cars benchmarks, due to the limitations of diffusion models in distinguishing fine-grained classes based on their class names. We further study when scaling the number of synthetic images in Appendix~\ref{app:sec:air_dtd_scale_up}. It shows that for \ours, the scaling curve meets a plateau at 500 images per class on DTD, while it keeps increasing over 5000 images per class on the Aircraft dataset.

\subsubsection{Comparison with Few-shot learning methods}
We compare our method with state-of-the-art methods in the few-shot learning literature. CoOp~\cite{coop} optimizes the learnable token of textual input, TIP-Adapter~\cite{tipadapter} designs a cache model from the few-shot training set, and AMU-Tuning~\cite{amutuning} balances the CLIP logit with MOCOv3~\cite{mocov3}.
We reproduce the results using the official source code of each. While the few-shot learning methods optimize a pre-trained model using real data only, with \ours method, we use both the few-shot real images as well as 500 synthetic images per class for the training set. CLIP ViT-B/16 is used as a base model, and we conduct experiments in the 16-shot setting. 

The results are presented in Table~\ref{tab:fewshot_sota_comparison}. Our \ours method outperforms baseline methods on 8 out of 9 benchmarks, achieving the best average accuracy of 87.2\% vs. 82.1\% of the next best method TIP-Adapter-f. We observe that there are significant performance gaps between \ours and baseline methods on the Aircraft and Cars datasets, which consist of fine-grained classes. This indicates the advantages of using the synthetic data of \ours in addition to the few-shot real images.
Qualitative examples on these datasets can be found in Appendix~\ref{app:sec:qual_finegrained}.

\subsection{Per-class analysis on ImageNet}
\label{subsec:analysis}

To identify how different factors in the synthetic dataset impact performance, following \citet{scalelaw}, we evaluate two metrics: recognizability and diversity. For our analysis, we randomly sample 50 images per ImageNet class for each synthetic dataset generation method.

\begin{itemize}
    \item Recognizability: To evaluate the fidelity of the generated images, we use a pre-trained ImageNet ViT-B/16 classifier (accuracy of 86.2\%) to classify the generated images. The F1 score for each class serves as the metric.
    \item Diversity: For each class, we extract features from the same pre-trained ImageNet ViT-B/16 classifier and compute their standard deviation as a measure of diversity in the generated images.
\end{itemize}
Figure~\ref{fig:per_class_analysis} presents the scatter plots for each method where each point summarizes one class. The color of each point indicates the log-likelihood of the corresponding class in the validation set of ImageNet, as predicted by the CLIP model fine-tuned on 0.5M synthetic images in the 16-shot setting.

\myparagraph{Recognizability and diversity are inversely correlated.} 
Across all methods, there is an inverse correlation between recognizability and diversity: as recognizability increases, diversity tends to decrease, and vice versa.

\myparagraph{Few-shot guided methods exhibit higher recognizability, while other methods have higher diversity.}
DataDream and LoFT show higher recognizability compared to ClassPrompt and CaptionPrompt. This is because these few-shot guided methods are specifically trained to generate images similar to the real images. This alignment improves the realism and quality of the generated images, leading to higher recognizability. While LoFT incorporates multiple LoRA weights to increase diversity, it still shows less diversity than ClassPrompt and CaptionPrompt.  

\myparagraph{Distinct strengths of each method.}
CaptionPrompt obtains a good performance on classes with high diversity (i.e., in the range of 20-25). On the other hand, DataDream and LoFT demonstrate better performance with high recognizability (i.e., when it is greater than 0.8). This suggests that each method has its own strengths. It remains an open question for further exploration of methods that combine these strengths, achieving high diversity and recognizability.

\subsection{Qualitative comparison}
\label{subsec:qual}

We present qualitative results in Figure~\ref{fig:qualitative_example} to gain insights into the diversity and quality of images from different methods. 
For the acoustic guitar class (Figure~\ref{fig:qual_sub1}), real images have high variety, including differences in zooming, color palettes, and the presence of humans.
In contrast, ClassPrompt images lack this diversity, displaying limited color variation and similar representations of the guitar.
DataDream demonstrates a better level of diversity, generating images with various colors and textures.
However, some DataDream images exhibit artifacts, which can detract from their overall quality.
In contrast, our \ours method successfully balances diversity and image quality.
This is reflected in the quantitative metrics, where \ours achieves the highest scores for recognizability (0.94) and diversity (14.68), outperforming ClassPrompt (0.88, and 9.54).

In the hourglass category in Figure~\ref{fig:qual_sub2}, ClassPrompt sometimes produces image related to ``hour'' but not ``hourglass'', as seen in the second column. Some of the images by DataDream show barely recognizable shapes of an hourglass. In contrast, images from \ours closely resemble the object of interest. This observation is consistent with the quantitative metrics where \ours achieves a recognizability score of 1.0 while ClassPrompt and DataDream score 0.89 and 0.95, respectively.

\begin{table}[t]
\centering
\setlength{\tabcolsep}{2.5pt}
\resizebox{0.47\textwidth}{!}{
\renewcommand{\arraystretch}{1.3}
\begin{tabular}{l|cccc}
\toprule
Fusing representation & 0.05M & 0.1M & 0.25M & 0.5M \\ \hline \hline
Caption embeddings & 70.36 & 70.55 & 70.62 & 70.66 \\
Image embeddings & 69.88 & 69.95 & 69.99 & 70.02 \\
Tokens from Textual Inversion & 69.70 & 70.13 & 70.21 & 70.29 \\
LoRA weights (= LoFT, ours) & \bf 71.02 & \bf 71.37 & \bf 72.05 & \bf 72.21 \\ 
\bottomrule
\end{tabular}
}
\vspace{-5pt}
\caption{Comparison of methods on fusing different representations when fine-tuning CLIP. Experiments are done in the 16-shot setting on ImageNet.
}
\label{tab:mixed_representations}
\vspace{-10pt}
\end{table}

\subsection{Ablation study of \ours}
\label{subsec:ablation}

Our ablation study investigates how different methods of fusing representations impact the synthetic training dataset. We conduct two studies: one exploring the effect of various fusion techniques, and another examining the influence of the weight parameter $\lambda$ in our LoRA-based fusion method. 

\subsubsection{Fusion on different representations}

To explore the effectiveness of different fusion techniques on the resulting synthetic dataset, we compare three methods for fusing representations: 1) caption embedding fusion, which involves averaging the text embeddings of the PaliGemma captions from two images;
2) image embedding fusion, which directly embeds two images using an image encoder, and then passes the average image representation to Stable-Diffusion-2.1-unclip\footnote{https://huggingface.co/stabilityai/stable-diffusion-2-1-unclip}, a model for image-to-image generation;
and 3) Textual Inversion~\cite{textual_inv} fusion, which optimizes input tokens for each image and then fuses the learned token embeddings from two images.

We generate up to 500 images per class on ImageNet using each method and fine-tune CLIP on the resulting datasets. Our results in the 16-shot setting are presented in Table~\ref{tab:mixed_representations}, which shows that our LoRA-based fusion method outperforms the other techniques both when generating fewer samples (50K imgs, +0.66\%) and with increasing number of generations (500K imgs, +1.55\%). This suggests that our method is more effective at capturing the underlying structure of the data and generating high-quality images.

\subsubsection{$\lambda$ variation for LoRA fusion}

We also examine the impact of $\lambda$ on the generated images and downstream classification performance. In \cref{sec:qualitative_lambda} we show examples of images generated with different values of $\lambda$ illustrating that $\lambda = 0.5$ provides the best visual results, especially in terms of diversity.

\setlength{\columnsep}{10pt}   
\begin{wraptable}{r}{0.22\textwidth}
\vspace{-12pt}
\centering
\setlength{\tabcolsep}{4pt}
\resizebox{0.22\textwidth}{!}{
\renewcommand{\arraystretch}{1.2}
\begin{tabular}{l|cc}
\toprule
Fusing LoRAs & 0.05M & 0.5M \\
\hline\hline
$\lambda = 0.5$ & 29.03 & \textbf{45.41} \\
$\lambda = 0.7 \,\,(\text{or } 0.3)$ & 25.60 & 39.18 \\
$\lambda = 1 \quad\,(\text{or } 0)$ & 22.42 & 30.85 \\
$\lambda \sim \text{Beta}(2,2)$ & 28.28 & 43.15 \\
$\lambda \sim \text{Beta}(5,5)$ & 28.56 & 44.91 \\
$\lambda \sim \text{Beta}(10,10)$ & \textbf{29.40} & 45.36 \\
$[0.5, 0.25, 0.25]$ & 27.87 & 43.63 \\
$[0.33, 0.33, 0.33]$ & 28.46 & 44.10 \\
$[0.7, 0.15, 0.15]$ & 22.19 & 36.28 \\
\bottomrule
\end{tabular}
}
\vspace{-10pt}
\caption{Ablation study on LoRA fusion. We train ResNet50 from scratch in the 16-shot setting.}
\label{table:ablation_loft_lora_combination}
\vspace{-10pt}
\end{wraptable}
In Table~\ref{table:ablation_loft_lora_combination}, we quantitatively demonstrate the importance of choosing an optimal value for $\lambda$. When $\lambda$ is set to 0.5, our method achieves the best performance, with a significant increase in accuracy as the amount of training data increases (scaling from 29.03\% to 45.41\% with 0.05M to 0.5M samples). In contrast, setting $\lambda$ to values closer to 0 or 1 results in lower performance, due to a lack of diversity in the generated images (only scaling from 22.42\% to 30.85\% with 0.05M to 0.5M samples for $\lambda = 1$).

Second, we explore the impact of introducing randomness to the $\lambda$ value, using a Beta distribution 
$\text{Beta}(\alpha, \alpha)$, where $\alpha$ controls the concentration of $\lambda$ around the value of 0.5. Larger values of $\alpha$ lead to a distribution that is more concentrated around 0.5, while smaller values of $\alpha$ allow for a broader spread across the [0,1] interval. When $\alpha$ is small, the performance decreases, i.e., accuracy of 43.15\% ($\alpha=2$) vs. 45.36\% ($\alpha=10$) with 0.5M data samples. This suggests that having a more concentrated distribution around $\lambda = 0.5$ is beneficial for performance.

Finally, we also evaluate the performance of fusing three LoRA weights during dataset generation. Our results, presented in the last three rows of Table~\ref{table:ablation_loft_lora_combination}, show that none of these methods outperform the two-LoRA fusion with $\lambda=0.5$. In fact, using more than two LoRAs introduces artifacts into the generated images, which can deteriorate their recognizability.

\section{Conclusion}

In this paper, we introduced \ours, \ourslong. \ours fine-tunes LoRA weights on individual real images, ensuring high fidelity when generating synthetic images, and then fuses them to achieve diversity. Our experiments demonstrate that \ours consistently outperforms other methods when fine-tuning a pre-trained CLIP model. This is because the synthetic images generated by \ours complement the prior knowledge contained in CLIP by accurately capturing and fusing the features of each class. Additionally, we showed that \ours performs better as the number of few-shot samples increases when training from scratch. Our analysis of the synthetic datasets showed that \ours achieves a balance of high fidelity with reasonable diversity, while methods like ClassPrompt and CaptionPrompt focus more on generating diverse images at the cost of fidelity.

\section*{Broader impact and limitations}

\myparagraph{Broader impact.} Our method, LoFT, presented a promising approach for generating high-quality synthetic data with few-shot guidance. This could reduce the cost associated with data collection and annotation. Our approach could be applied to a wide range of domains, including medical imaging, autonomous driving, and remote sensing, where acquiring large amounts of labeled data is challenging.

\myparagraph{Limitations.} Despite its advantages, LoFT does come with some limitations. A key challenge is the increased number of LoRA weights required in the few-shot setting. In the 16-shot scenario, LoFT needs 16 times more LoRA weights than methods like DataDream, which can lead to considerable memory and storage overhead. To mitigate this, we used a smaller LoRA rank (rank=2), which was sufficient to achieve better performance than DataDream (rank=16), while minimizing computational costs and storage requirements. However, the computational expense of fine-tuning these additional weights could be a limiting factor, particularly in tasks with many real samples as guidance. 

Additionally, the quality of the synthetic data generated by LoFT depends heavily on the quality of the few-shot data provided. If the few-shot samples are biased or do not adequately represent the broader distribution, the synthetic images may fail to capture the diversity. Mitigating the bias to ensure a more diverse representation of the data would be an interesting direction for future work.

\section*{Acknowledgements}

Jae Myung Kim thanks the International Max Planck Research School for Intelligent Systems (IMPRS-IS)
and the European Laboratory for Learning and Intelligent Systems (ELLIS) PhD programs for support. 
This work was partially funded by the ERC (853489 - DEXIM) and the Alfried Krupp von Bohlen und Halbach Foundation, which we thank for their generous support. Cordelia Schmid
would like to acknowledge the support by the Körber Euro-
pean Science Prize. The authors gratefully acknowledge the Gauss Centre for Supercomputing e.V. (www.gauss-centre.eu) for funding this project by providing computing time on the GCS Supercomputer JUWELS at Jülich Supercomputing Centre (JSC).



{
    \small
    \bibliographystyle{ieeenat_fullname}
    \bibliography{main}
}

\clearpage
\appendix
\twocolumn[
\section*{\centering Supplementary Material for \\LoFT: LoRA-Fused Training Dataset Generation with Few-shot Guidance}
\vspace{30pt}
]

\renewcommand\thesection{\Alph{section}}

\section{Implementation details of classifier training}

When fine-tuning the CLIP model, we freeze the CLIP parameters and update the LoRA weights applied to them, with a lora rank of 16. We use a batch size of 256 and a learning rate of 1e-6 with cosine annealing for the learning rate schedule. The weight decay is set to 1e-4. As the dataset size increases, we adjust the number of iterations to be increased while decreasing the number of epochs. For dataset sizes of 0.05M, 0.1M, 0.25M, 0.5M, and 1M, the number of epochs is set to 90, 80, 70, 60 and 50, respectively. The warm-up period is set to 10\% of the total epochs, resulting in 9, 8, 7, 6, and 5 warm-up epochs for each dataset size.

When training the ResNet50 from scratch, we use a batch size of 2048 and a learning rate of 0.2 with cosine annealing for the learning rate schedule. The weight decay is set to 1e-4. As the dataset size increases, we adjust the number of iterations to be increased while decreasing the number of epochs. For dataset sizes of 0.05M, 0.1M, 0.25M, 0.5M, and 1M, the number of epochs is set to 300, 250, 200, 150, and 100, respectively. The warm-up period is set to 10\% of the total epochs, resulting in 30, 25, 20, 15, and 10 warm-up epochs for each dataset size.

\section{Training an image classifier from scratch}
\label{app:sec:exp_imagenet_rn50}

\begin{figure*}[t]
    \centering
    \includegraphics[width=\linewidth]{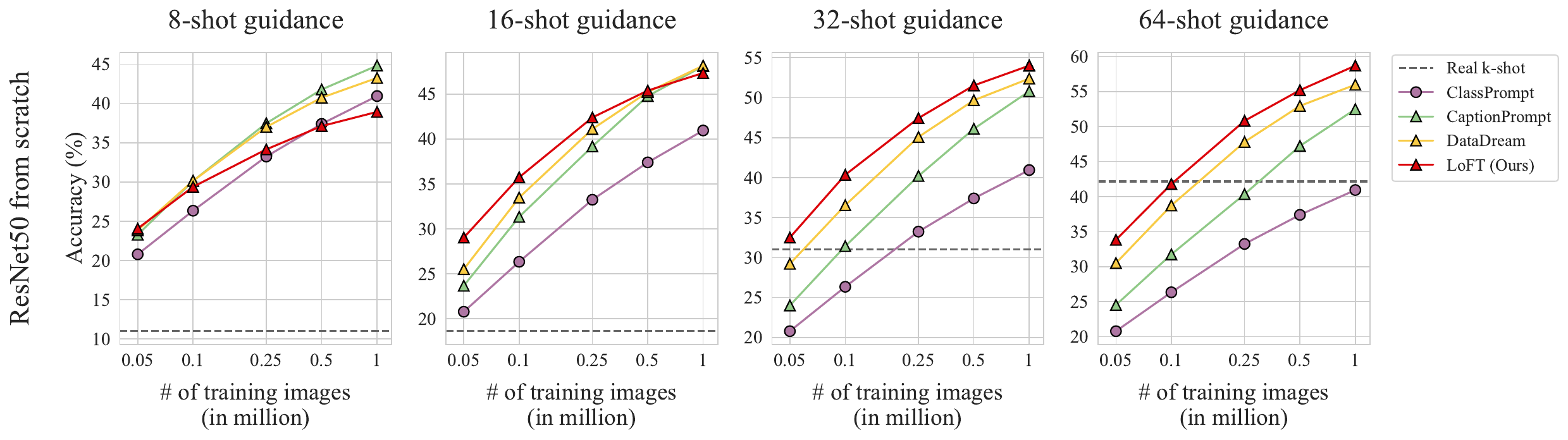}
    \vspace{-15pt}
    \caption{
    Classification accuracy on ImageNet when training ResNet50 from scratch on synthetic data generated from different methods at different scales. We report few-shot guidance on 8, 16, 32, and 64 images per class and a baseline of training CLIP only on k-shot real data. 
    \ours benefits from a larger number of real images as guidance. 
    }
    \label{fig:main_result_resnet50}
\end{figure*}

We further evaluate the dataset generation methods by training ResNet50~\cite{resnet} model from scratch and evaluating it on the validation dataset of ImageNet.
The results are shown in Figure~\ref{fig:main_result_resnet50}.

\myparagraph{Performance improves with dataset size.}
Contrary to CLIP fine-tuning, training from scratch with data from the ClassPrompt method improves with increasing dataset size. Similarly, the performance of all few-shot guided generation methods also improves consistently with data scale for all k-shot settings.

\myparagraph{The best method depends on the k-shot setting.}
The best method differs depending on the number of k-shot real images used for guidance. In the case of smaller k-shot settings (8-shot), CaptionPrompt outperforms both DataDream and LoFT. However, as the number of shots increases, DataDream and LoFT start to outperform CaptionPrompt, with LoFT consistently performing better than DataDream. For instance, LoFT achieves 58.70\% at 1M scale in 64-shot while DataDream and CaptionPrompt achieve 56.00\% and 52.48\%, respectively. 

\myparagraph{Conclusions differ between fine-tuning and training from scratch.}
While the best method for training from scratch depends on the number of k-shot real images used for guidance, LoFT outperforms all baseline methods consistently across all k-shot settings when fine-tuning the CLIP model. 
Synthetic images by \ours complement the prior knowledge contained in CLIP because it generates images
that accurately represent the features of each class.
In contrast, CaptionPrompt introduces greater diversity
but since CLIP has already been pre-trained on a broad range of diverse images,
it does not provide as much complementary value, limiting its effectiveness for fine-tuning.

\begin{figure}
    \centering
    \includegraphics[width=\linewidth]{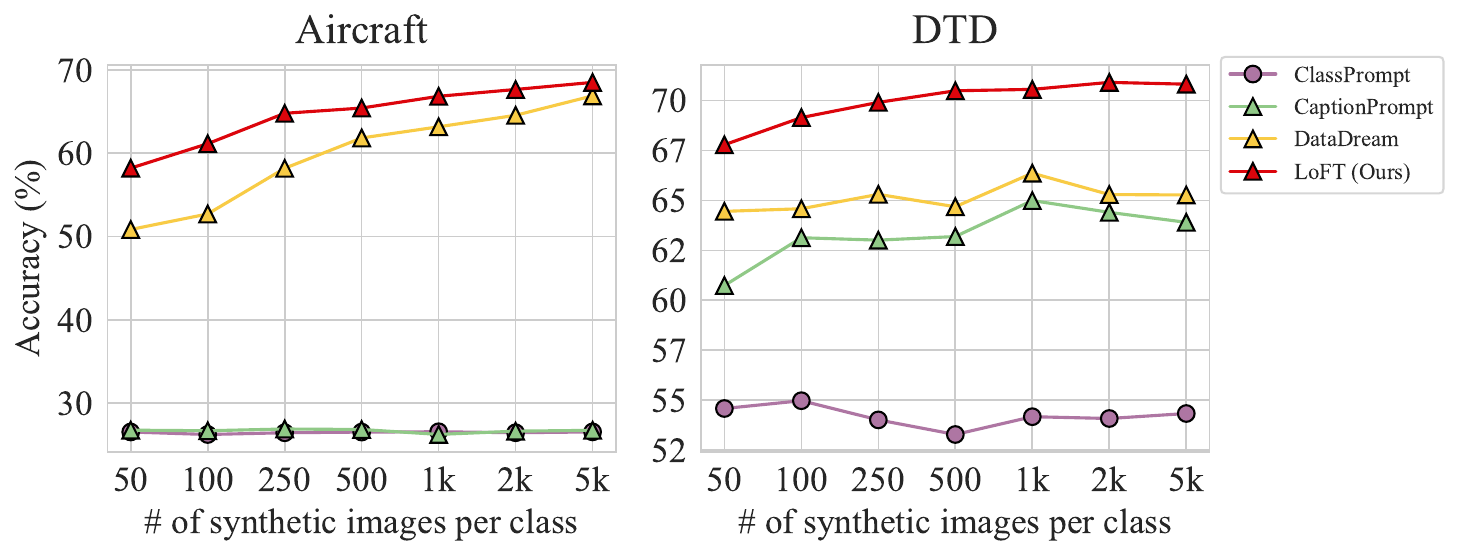}
    \caption{Scaling the number of synthetic data on Aircraft and DTD datasets when fine-tuning CLIP.}
    \label{app:fig:scaleup}
\end{figure}

\section{Scaling up to 5000 images per class on fine-grained datasets}
\label{app:sec:air_dtd_scale_up}
To study the scaling ability of synthetic data size on fine-grained dataset, we conduct experiments by generating up to 5000 images per class for the Aircraft and DTD datasets. For the few-shot synthetic data generation methods, we use 16-shot real images as guidance. The results from fine-tuning CLIP are shown in Figure~\ref{app:fig:scaleup}. ClassPrompt and CaptionPrompt reach a plateau from the beginning, indicating that increasing the number of synthetic images does not improve the classification performance. On DTD, \ours reaches a plateau at 500 images per class, while on the Aircraft, performance continues to improve up to 5000 images per class. 

\section{Qualitative results on fine-grained datasets}
\label{app:sec:qual_finegrained}

\begin{figure*}[t]
    \centering
    \begin{subfigure}{.5\textwidth}
      \centering
      \includegraphics[width=0.9\linewidth]{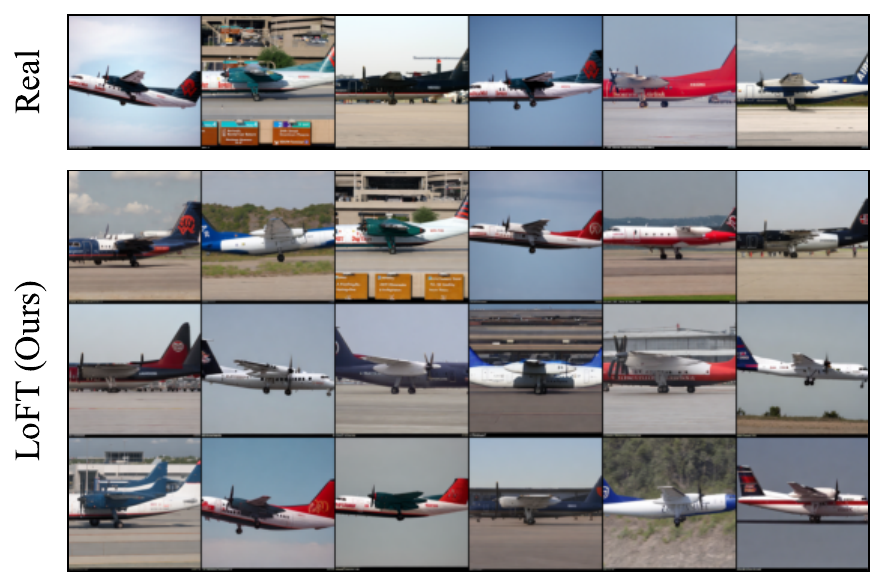}
      \caption{``DHC-8-100'' class on Aircraft}
      \label{fig:supp_aircraft_0}
    \end{subfigure}%
    \begin{subfigure}{.5\textwidth}
      \centering
      \includegraphics[width=0.9\linewidth]{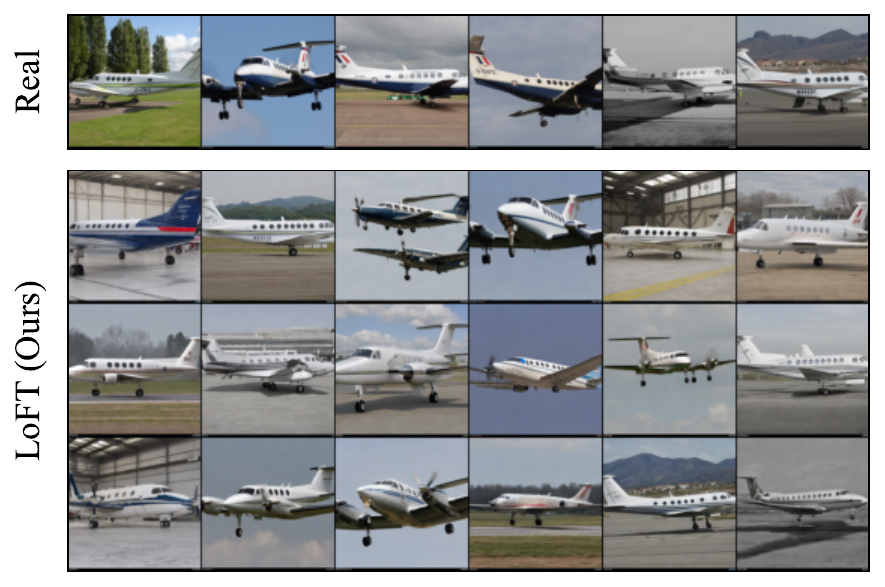}
      \caption{``Model B200'' class on Aircraft}
      \label{fig:supp_aircraft_1}
    \end{subfigure}
    
    \vspace{40pt}
    
    \begin{subfigure}{.5\textwidth}
      \centering
      \includegraphics[width=0.9\linewidth]{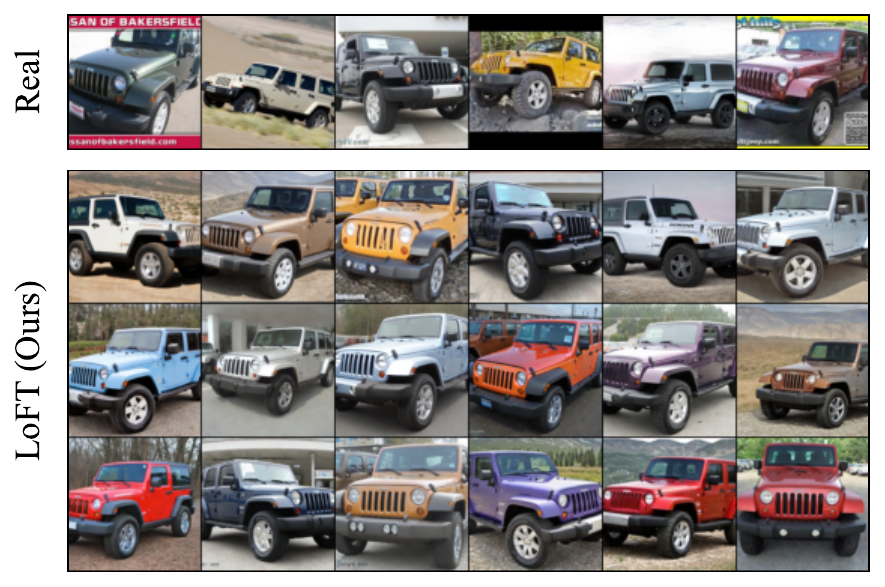}
      \caption{``Jeep Wrangler SUV 2012'' class on Cars}
      \label{fig:supp_cars_0}
    \end{subfigure}%
    \begin{subfigure}{.5\textwidth}
      \centering
      \includegraphics[width=0.9\linewidth]{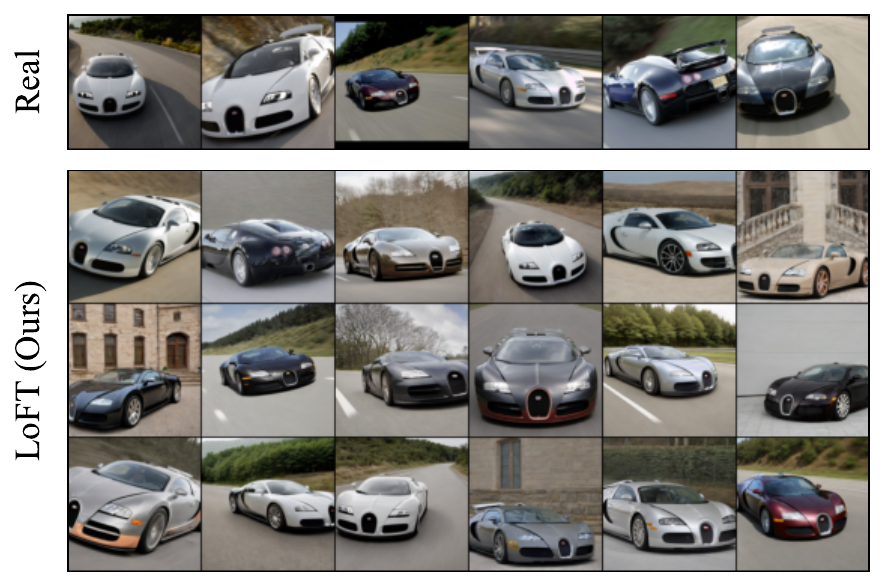}
      \caption{``Bugatti Veyron 16.4 Coupe 2009'' class on Cars}
      \label{fig:supp_cars_1}
    \end{subfigure}
    \vspace{10pt}
    \caption{Qualitative results of our \ours method on Aircraft and Cars datasets.} 
    \label{fig:supp_qual_finegrained}
\end{figure*}

We show qualitative results of our \ours method on Aircraft and Cars datasets. 
For the DHC-8-100 class on the Aircraft dataset in Figure~\ref{fig:supp_aircraft_0}, \ours generate a propeller attached to the wing, which resembles real images. Moreover, for the Model B200 class in Figure~\ref{fig:supp_aircraft_1}, \ours generates the shape of the class similar to real images, such as the head shape and the tail shape. Similarly in Figure~\ref{fig:supp_cars_0} and Figure~\ref{fig:supp_cars_1}, \ours generate images of the class ``Jeep Wrangler SUV 2012'' and ``Bugatti Veyron 16.4 Coupe 2009'' that resemble the shape and fine-grained details of real images, respectively.

\section{Additional per-class analysis}

\myparagraph{Correlation between diversity and alignment.}
In addition to the recognizability and diversity metrics introduced in \S\ref{subsec:analysis}, we introduce one additional metric, \textbf{alignment}, to measure how closely the distribution of synthetic data aligns with that of real data. To quantify this, we calculate the Fréchet Inception Distance (FID)~\cite{fid} score for each class, where a lower score indicates closer alignment between the synthetic and real data distributions.

\begin{figure*}
    \centering
    \includegraphics[width=\linewidth]{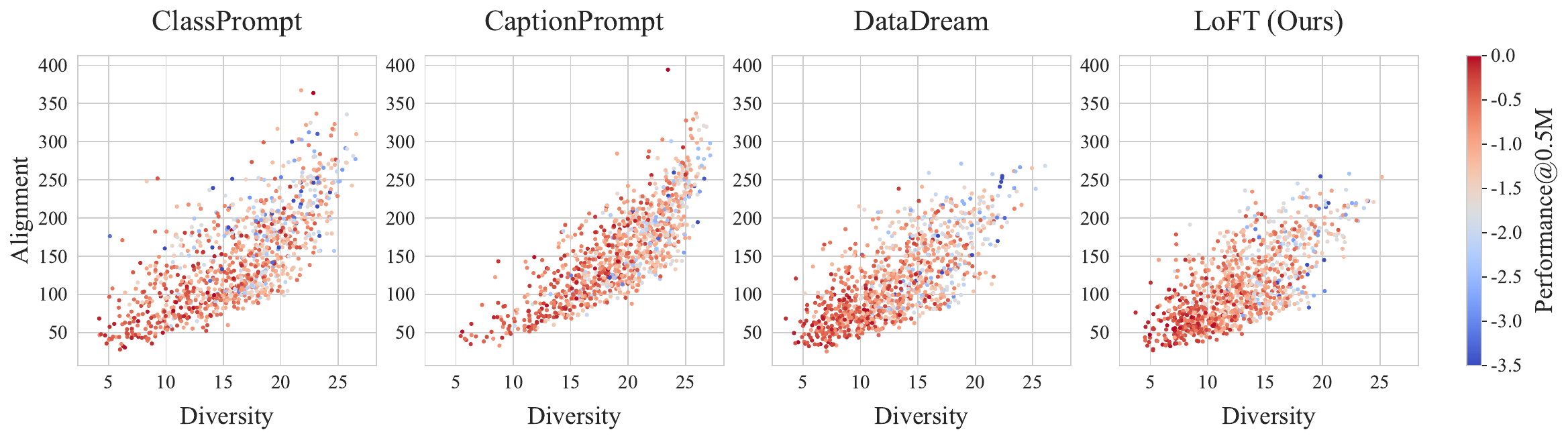}
    \vspace{-15pt}
    \caption{
    Per-class analysis of diversity and alignment in synthetic datasets generated from different methods. The color indicates a log-likelihood of the ImageNet validation dataset when CLIP is fine-tuned on the 0.5M-sized synthetic dataset in the 16-shot setting.
    }
    \label{fig:supp_per_class_analysis_diver_fid}
\end{figure*}

As seen in Figure~\ref{fig:supp_per_class_analysis_diver_fid}, we observe a positive correlation between alignment and diversity for all methods. This suggests that higher diversity in the generated images come at the expense of closely mimicking the real data distribution. Moreover, the overall alignment scores are smaller for LoFT and DataDream compared to ClassPrompt and CaptionPrompt, indicating that the images generated by LoFT and DataDream align more closely with the real data distribution.

\section{Ratio of data points the two models disagree on the prediction}

\begingroup
Since different methods show distinct strengths that contribute to performance gains, a natural question arises: do the classification models trained on each synthetic dataset exhibit different sets of corrected data points? To explore this, we use a ResNet50 model trained on the 0.5M-sized dataset from the ClassPrompt and 16-shot guided generation methods. We calculate the ratio of the number of data points in the validation ImageNet dataset that show inconsistent predictions relative to the total number of data points, i.e. where one model makes a correct prediction while the other model makes an incorrect one.
\setlength{\intextsep}{0pt} 
\setlength{\columnsep}{7pt} 
\begin{wrapfigure}{r}{0.22\textwidth} 
    \centering
    \includegraphics[width=\linewidth]{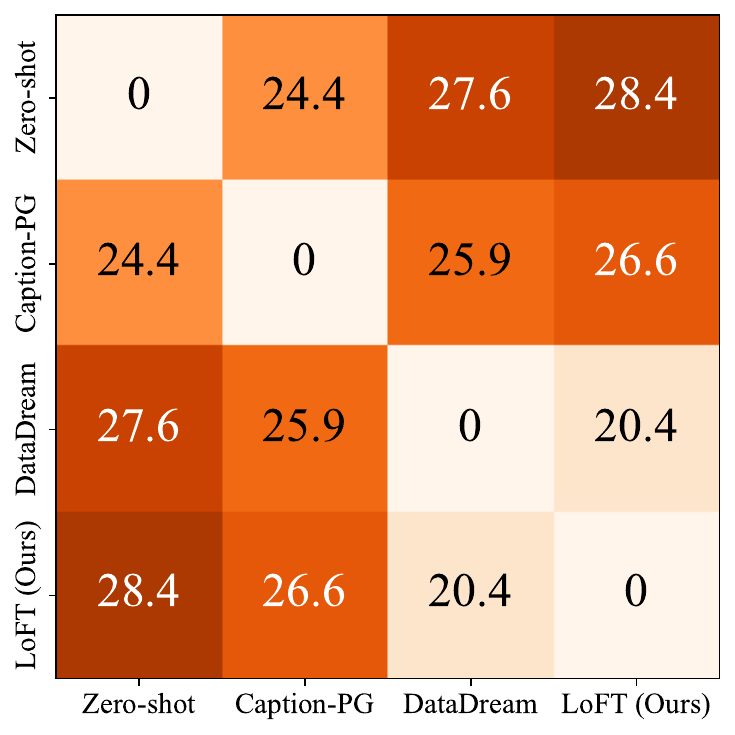}
    \vspace{-20pt}
    \caption{Ratio of data points the two models disagree on the prediction.}
    \label{fig:correction_flip_ratio}
\end{wrapfigure}
The results are shown in Figure~\ref{fig:correction_flip_ratio}. Even though the three few-shot guided methods (CaptionPrompt, DataDream, and LoFT) have comparable overall accuracy (around 45\% accuracy, in Figure~\ref{fig:main_result_resnet50}), the correction flip ratios between them are above 20\%. This suggests that each synthetic dataset encourages the model to learn different features. Moreover, LoFT shows a higher flip ratio with CaptionPrompt (26.6\%) compared to DataDream (20.4\%). This aligns with our per-class analysis, where CaptionPrompt maintains performance by leveraging greater diversity in image distribution, while LoFT and DataDream have higher recognizability, focusing more on image fidelity.

\endgroup

\section{Qualitative comparison on ImageNet}
\label{sec:qualitative_lambda}

\begin{figure*}[t]
    \centering
    \begin{subfigure}{.5\textwidth}
      \centering
      \includegraphics[width=0.95\linewidth]{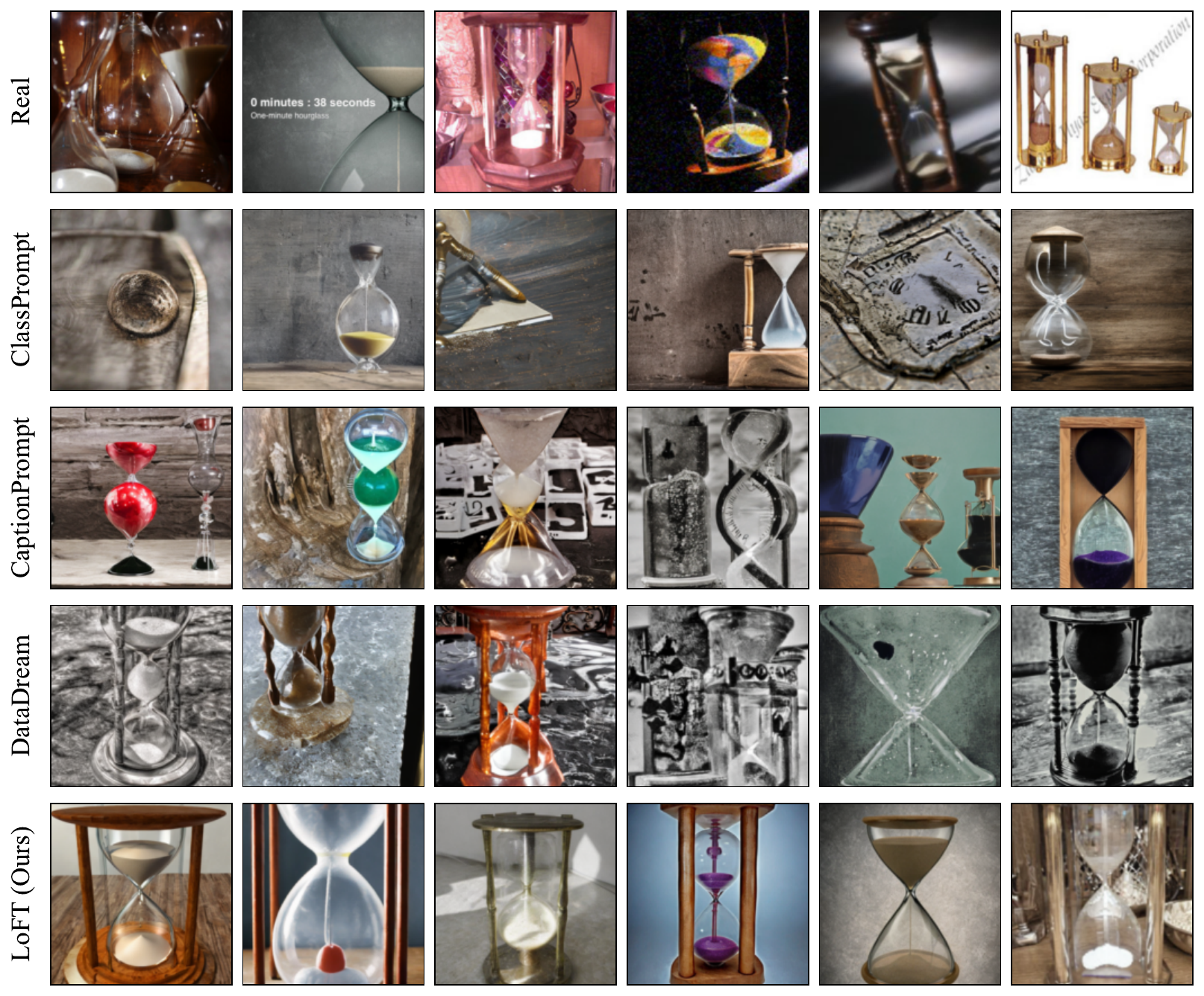}
      \caption{Hourglass}
      \label{fig:supp_qual_sub1}
    \end{subfigure}%
    \begin{subfigure}{.5\textwidth}
      \centering
      \includegraphics[width=0.95\linewidth]{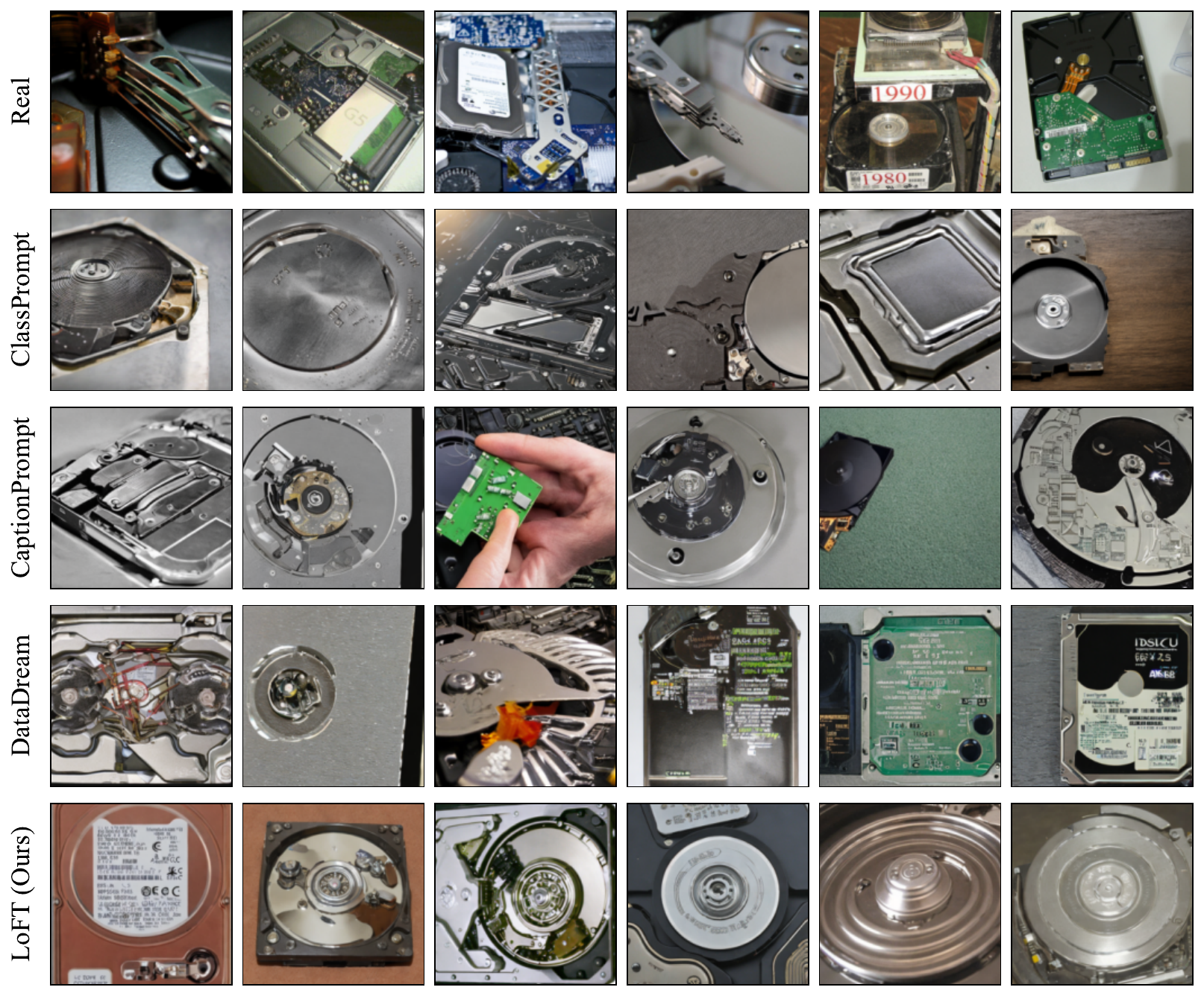}
      \caption{Hard disk drive}
      \label{fig:supp_qual_sub2}
    \end{subfigure}
    
    \vspace{40pt}
    
    \begin{subfigure}{.5\textwidth}
      \centering
      \includegraphics[width=0.95\linewidth]{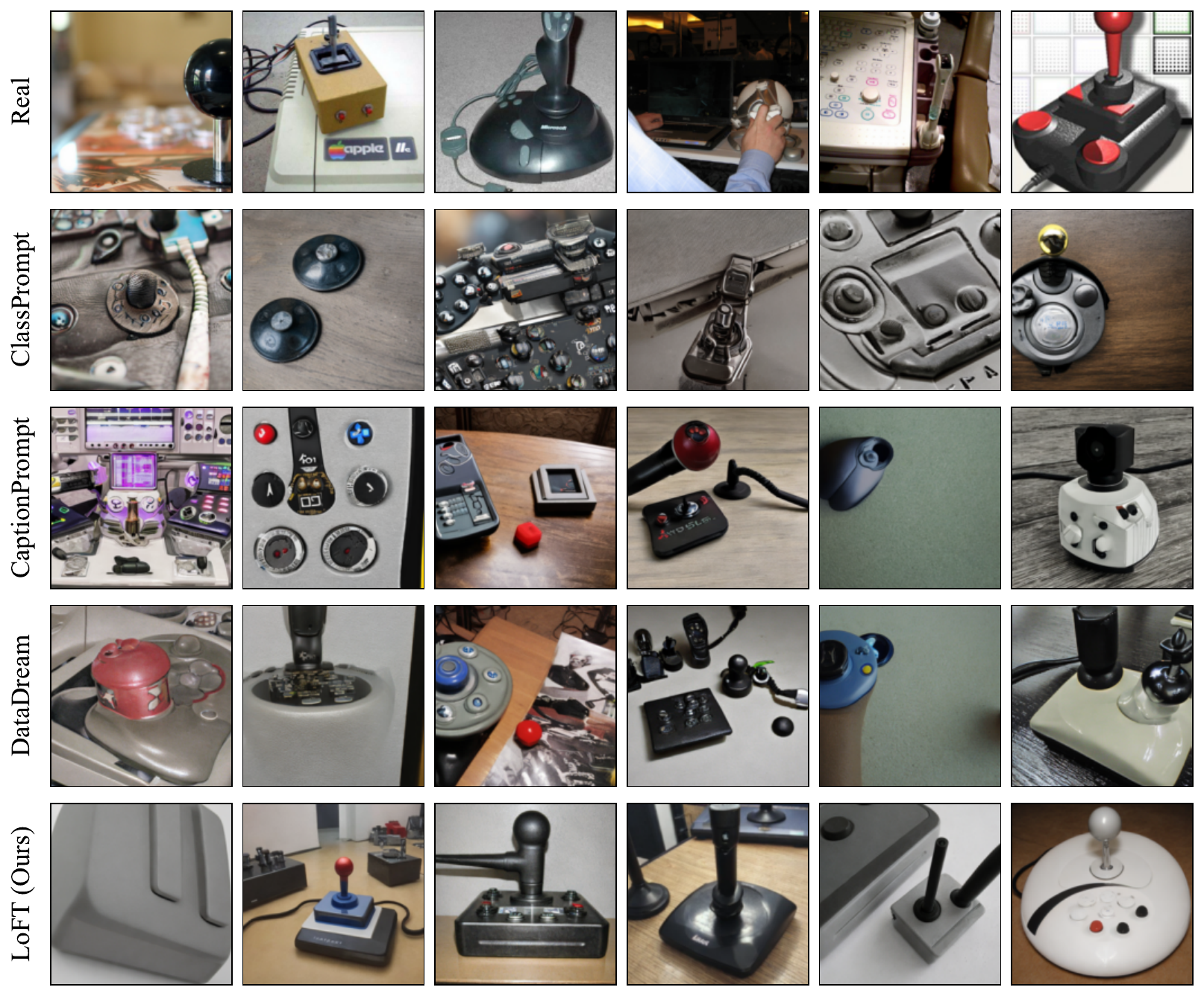}
      \caption{Joystick}
      \label{fig:supp_qual_sub3}
    \end{subfigure}%
    \begin{subfigure}{.5\textwidth}
      \centering
      \includegraphics[width=0.95\linewidth]{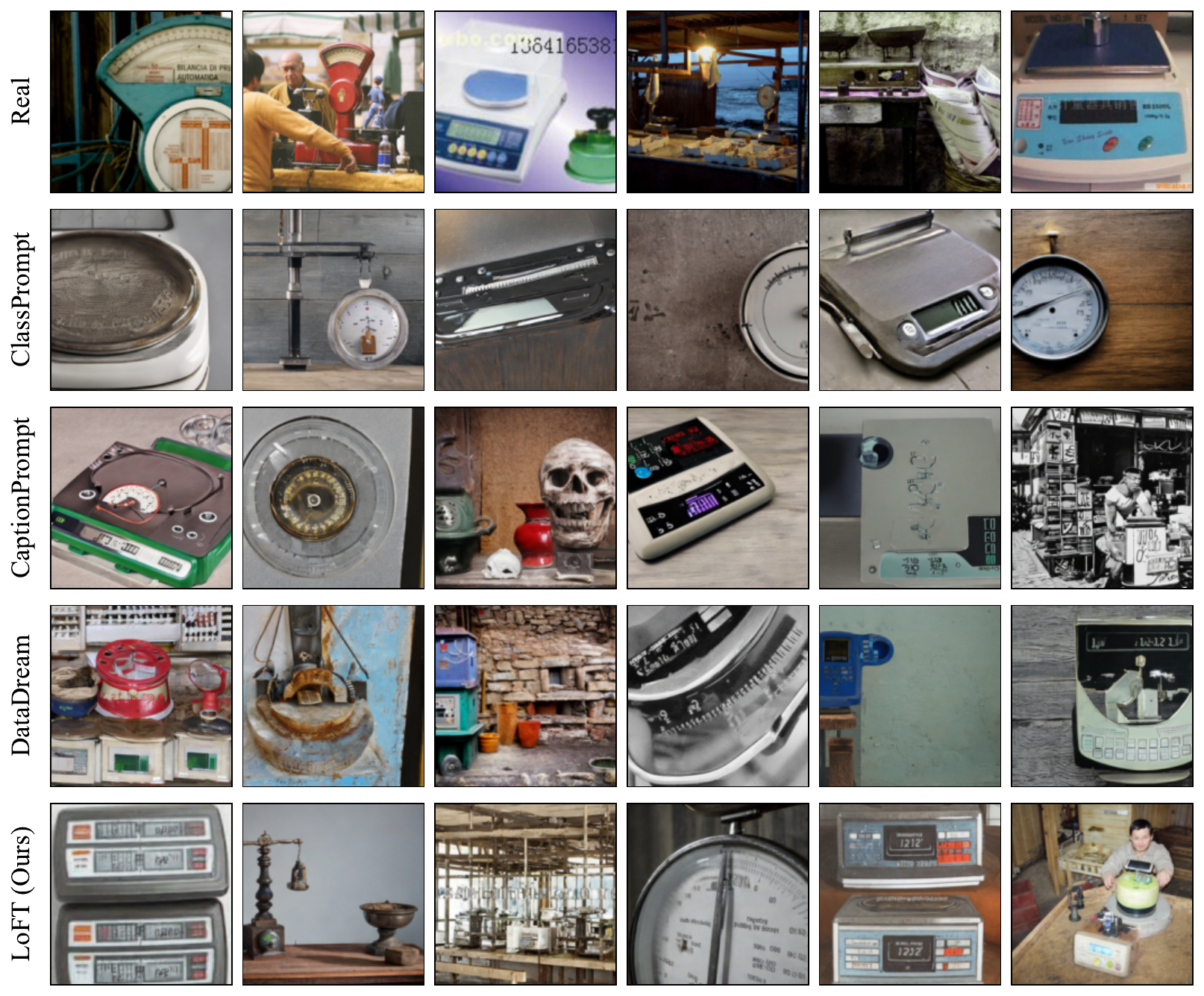}
      \caption{Weighing scale}
      \label{fig:supp_qual_sub4}
    \end{subfigure}
    \vspace{10pt}
    \caption{Qualitative examples for the classes Hourglass, Hard disk drive, Joystick, and Weighing scale.} 
    \label{fig:supp_qual_1}
\end{figure*}

\begin{figure*}[t]
    \centering
    \begin{subfigure}{.5\textwidth}
      \centering
      \includegraphics[width=0.95\linewidth]{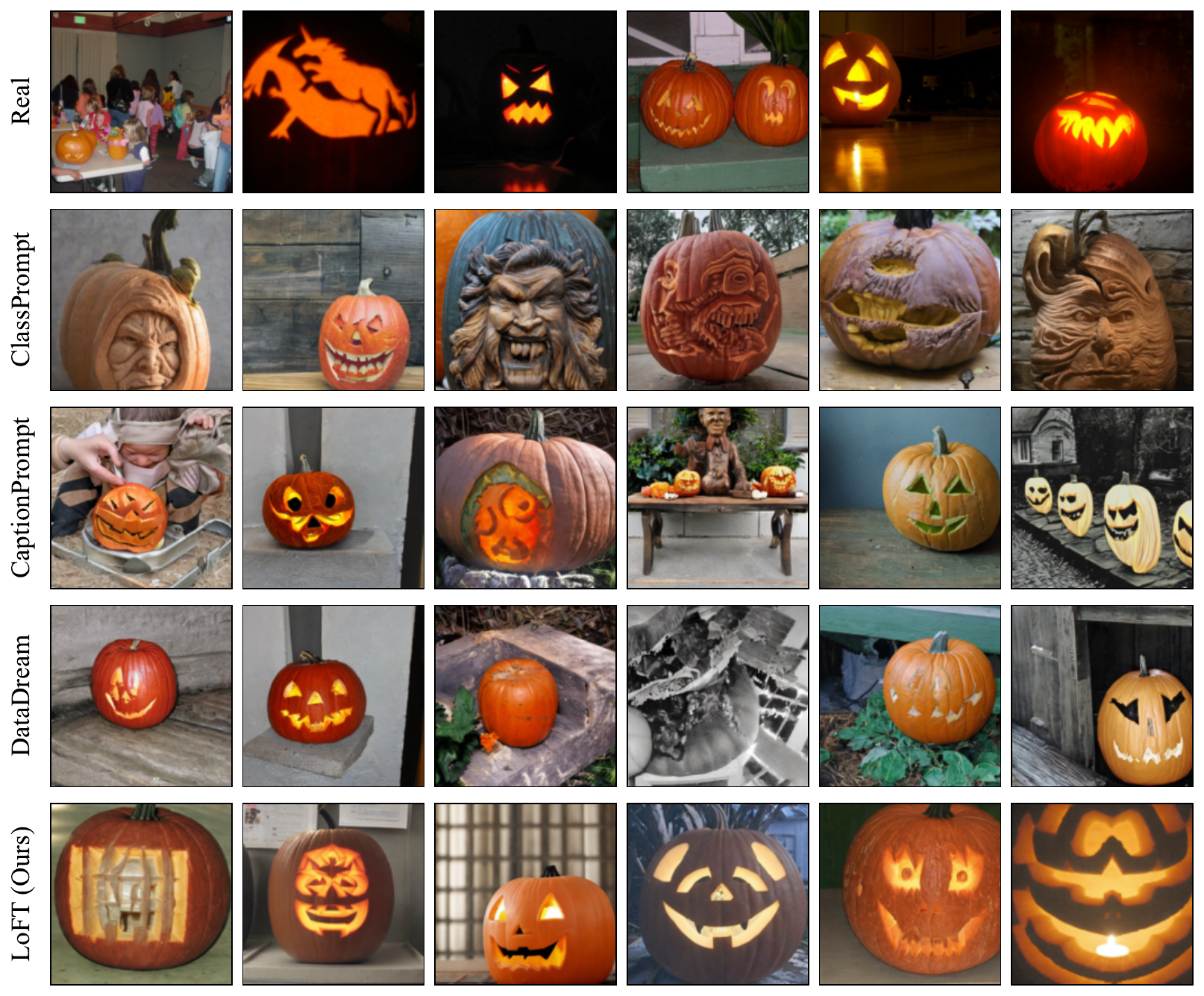}
      \caption{Carved Pumpkin}
      \label{fig:supp_qual_sub5}
    \end{subfigure}%
    \begin{subfigure}{.5\textwidth}
      \centering
      \includegraphics[width=0.95\linewidth]{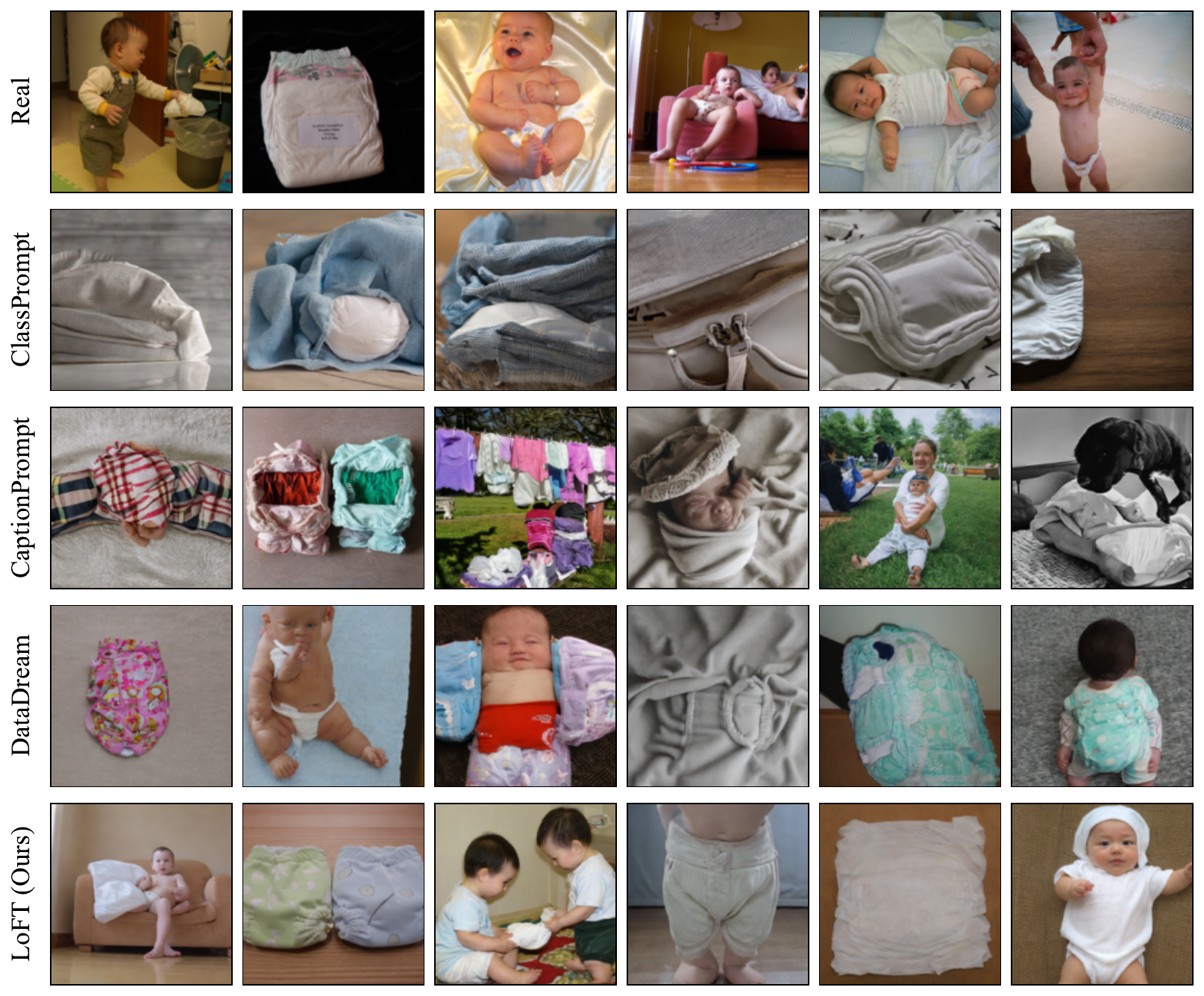}
      \caption{Diaper}
      \label{fig:supp_qual_sub6}
    \end{subfigure}
    
    \vspace{40pt}
    
    \begin{subfigure}{.5\textwidth}
      \centering
      \includegraphics[width=0.95\linewidth]{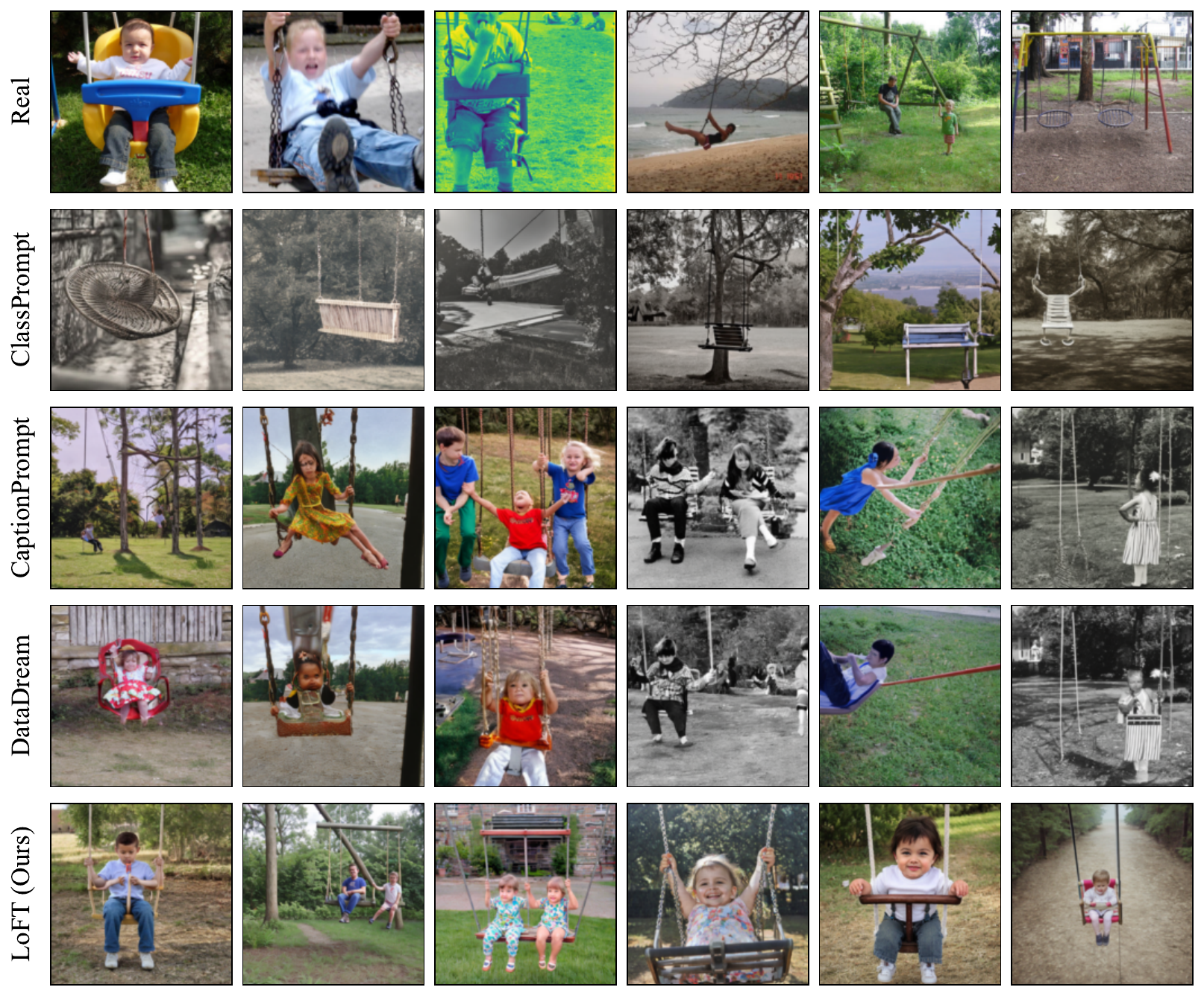}
      \caption{Swing}
      \label{fig:supp_qual_sub7}
    \end{subfigure}%
    \begin{subfigure}{.5\textwidth}
      \centering
      \includegraphics[width=0.95\linewidth]{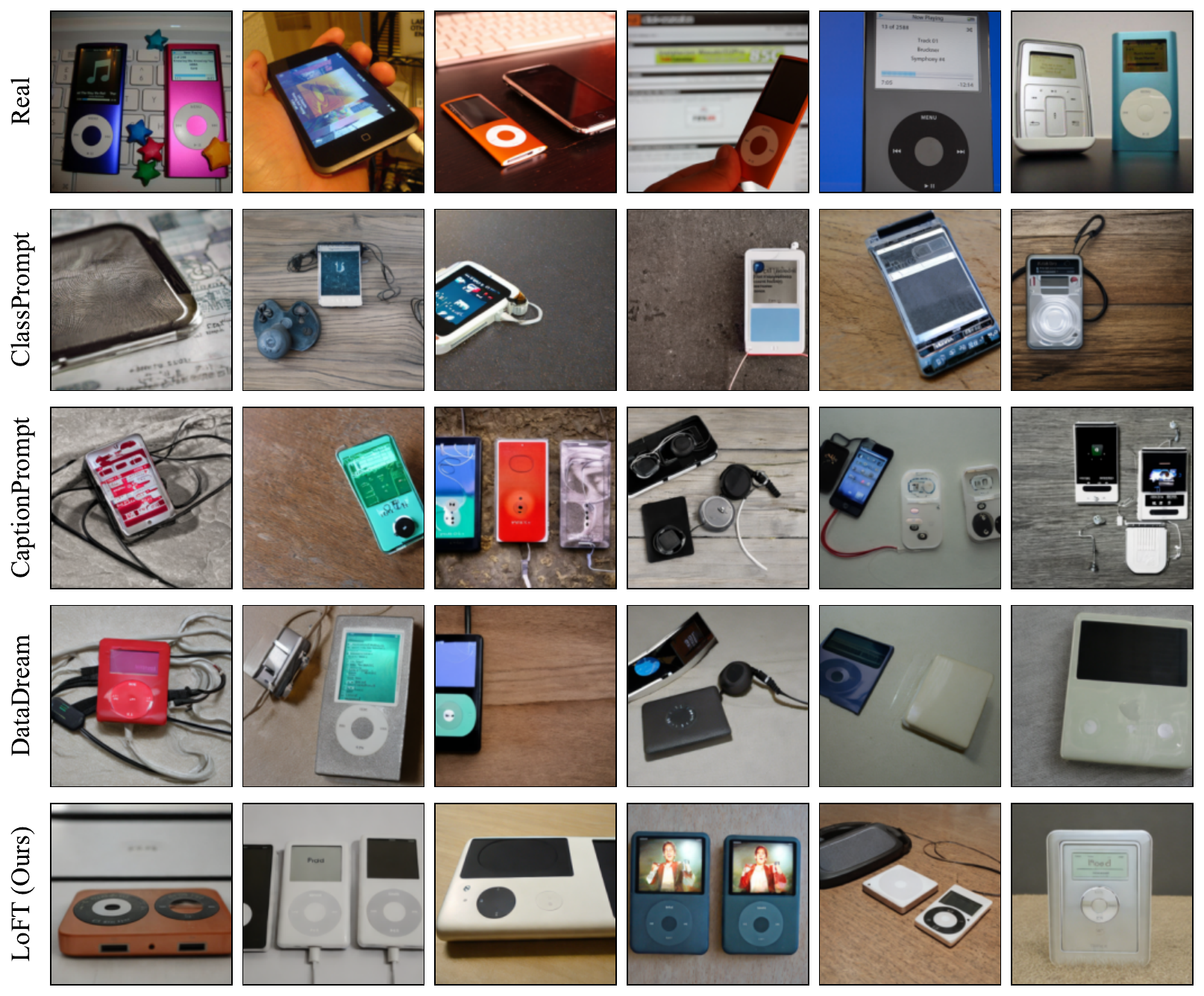}
      \caption{iPod}
      \label{fig:supp_qual_sub8}
    \end{subfigure}
    \vspace{10pt}
    \caption{Qualitative examples for the classes Carved Pumpkin, Diaper, Swing, and iPod.} 
    \label{fig:supp_qual_2}
\end{figure*}

We present additional qualitative results for 8 classes, i.e. Hourglass, Hard disk drive, Joystick, Weighing scale, Carved Pumpkin, Diaper, Swing, and iPod, in Figure~\ref{fig:supp_qual_1} and Figure~\ref{fig:supp_qual_2}. 

Taking the class Hourglass in Figure~\ref{fig:supp_qual_sub1} as an example, real images show hourglasses with diverse frames and varying sand colors. The images generated by ClassPrompt show less color variation. While CaptionPrompt and DataDream generate more colorful images, some of them are not easily recognizable as hourglasses. In contrast, LoFT generates images that maintain both diversity in the frame and sand color while clearly representing the hourglass.

Taking the class Swing in Figure~\ref{fig:supp_qual_sub7} as another example, real images show one or multiple swings, sometimes with a person riding them. Some of the generated images by ClassPrompt does not look like a swing, but rather resemble a chair. For CaptionPrompt and DataDream, some of the generated images focus more on the human subject than the swing itself, making the swing less visible. In contrast, LoFT generates clear images of swings or multiple swings, with the object clearly identifiable.

\section{Additional qualitative results varying $\lambda$}

\begin{figure}
    \centering
    \includegraphics[width=\linewidth]{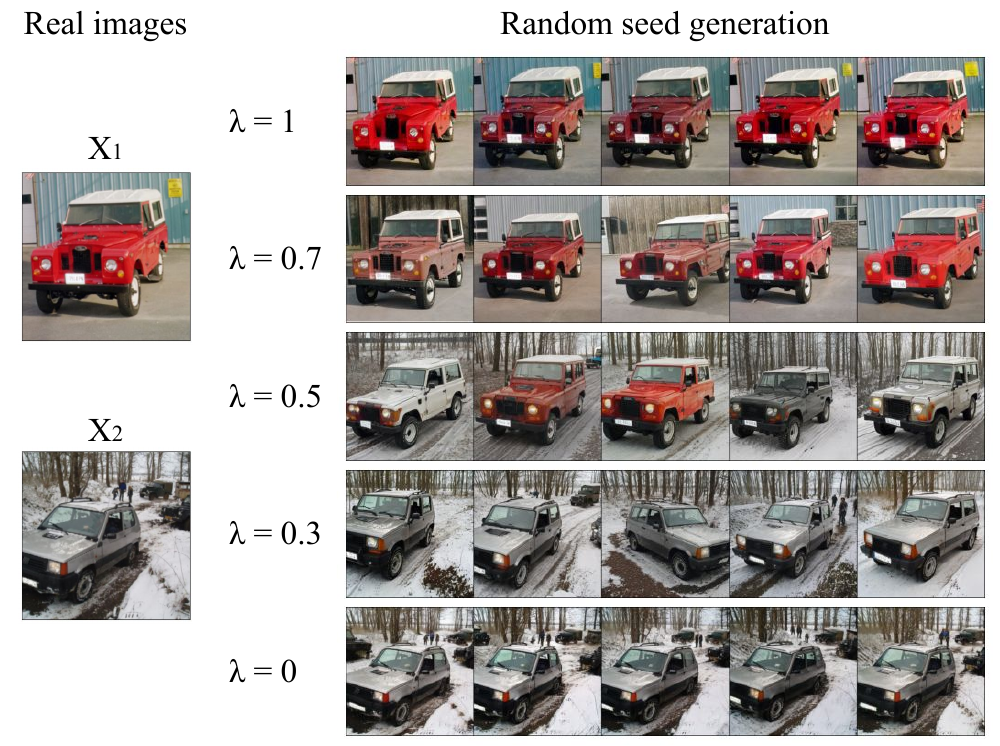}
    \caption{Ablation study of qualitative results on $\lambda$ variation when fusing LoRAs. Given two images of Jeep class, $\lambda=0.5$ merges features from both real images while maintaining diversity with random seed image generation. As $\lambda$ approaches 0 or 1, the generated images become more similar to the original image and looses diversity.}
    \label{fig:lambda_variation}
    \vspace{10pt}
\end{figure}

Figure~\ref{fig:lambda_variation} presents examples of images generated by our LoFT method with different $\lambda$ values, alongside their corresponding real images. As we adjust the weight parameter $\lambda$ for the LoRA fusion, we observe distinct trends in the generated outputs. When $\lambda$ is set to either 0 or 1, the generated images closely resemble the original real images. However, this approach limits the diversity of outputs across different seeds. As $\lambda$ approaches 0.5, we achieve an optimal balance that enhances the diversity of the generated images while preserving their quality. Each generated image effectively integrates features from the two original real images while resembling in-distribution data. This characteristic makes the synthetic training dataset produced by \ours beneficial for classification tasks.

\begin{figure*}[t]
    \centering
    \begin{subfigure}{.5\textwidth}
      \centering
      \includegraphics[width=0.9\linewidth]{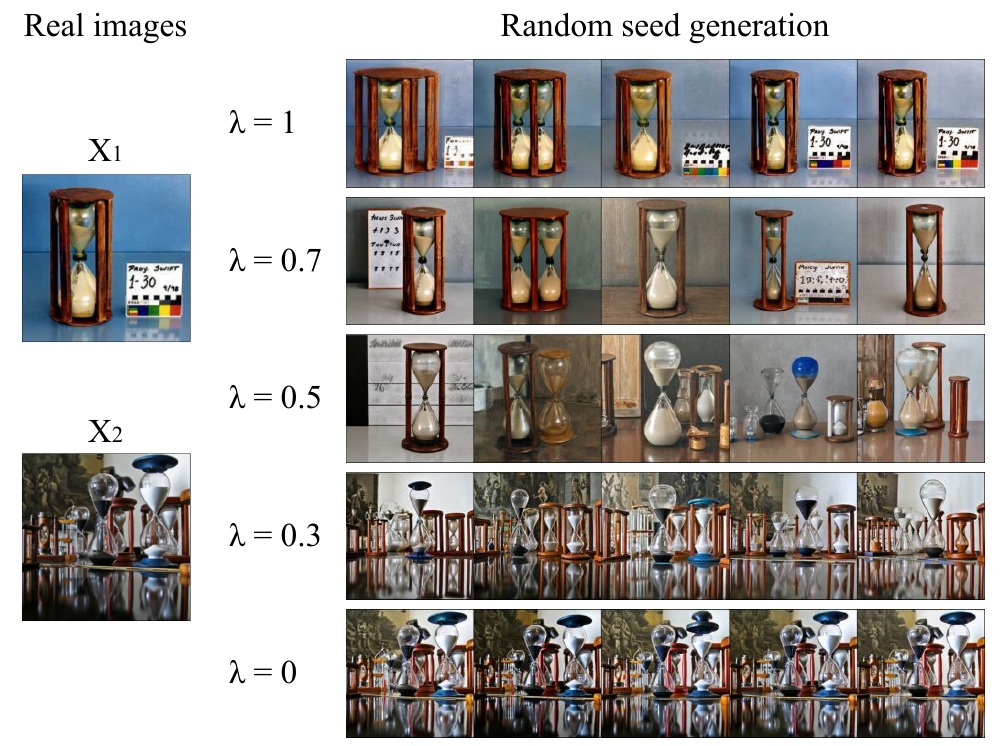}
      \caption{Hourglass}
      \label{fig:supp_lambda_sub1}
    \end{subfigure}%
    \begin{subfigure}{.5\textwidth}
      \centering
      \includegraphics[width=0.9\linewidth]{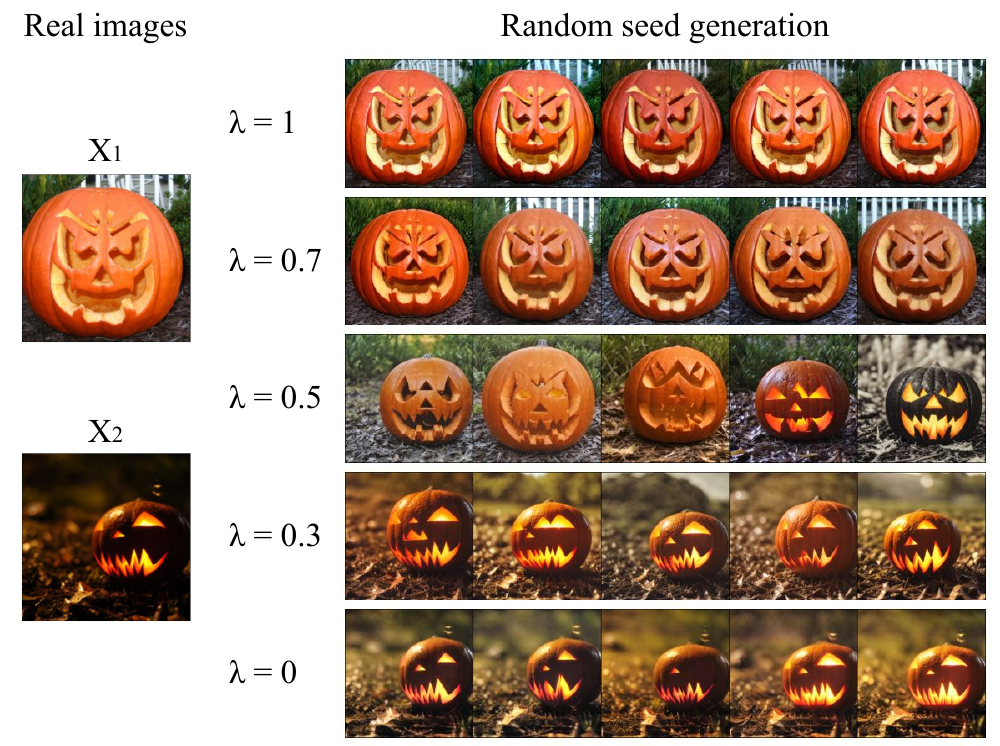}
      \caption{Carved pumpkin}
      \label{fig:supp_lambda_sub2}
    \end{subfigure}
    
    \vspace{40pt}
    
    \begin{subfigure}{.5\textwidth}
      \centering
      \includegraphics[width=0.9\linewidth]{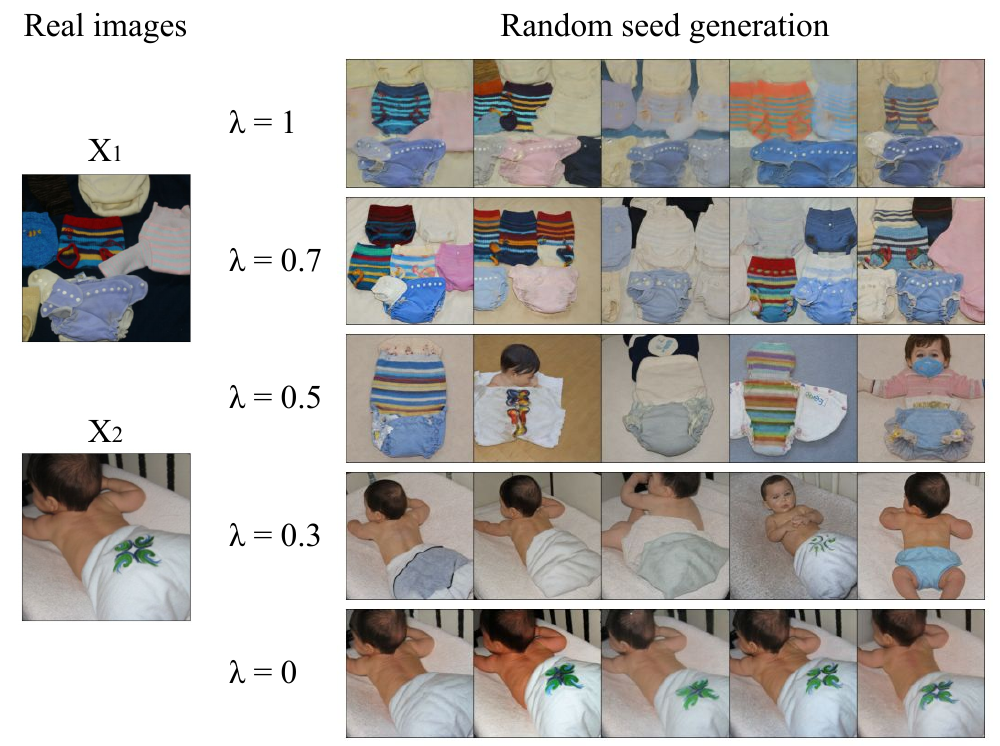}
      \caption{Diaper}
      \label{fig:supp_lambda_sub3}
    \end{subfigure}%
    \begin{subfigure}{.5\textwidth}
      \centering
      \includegraphics[width=0.9\linewidth]{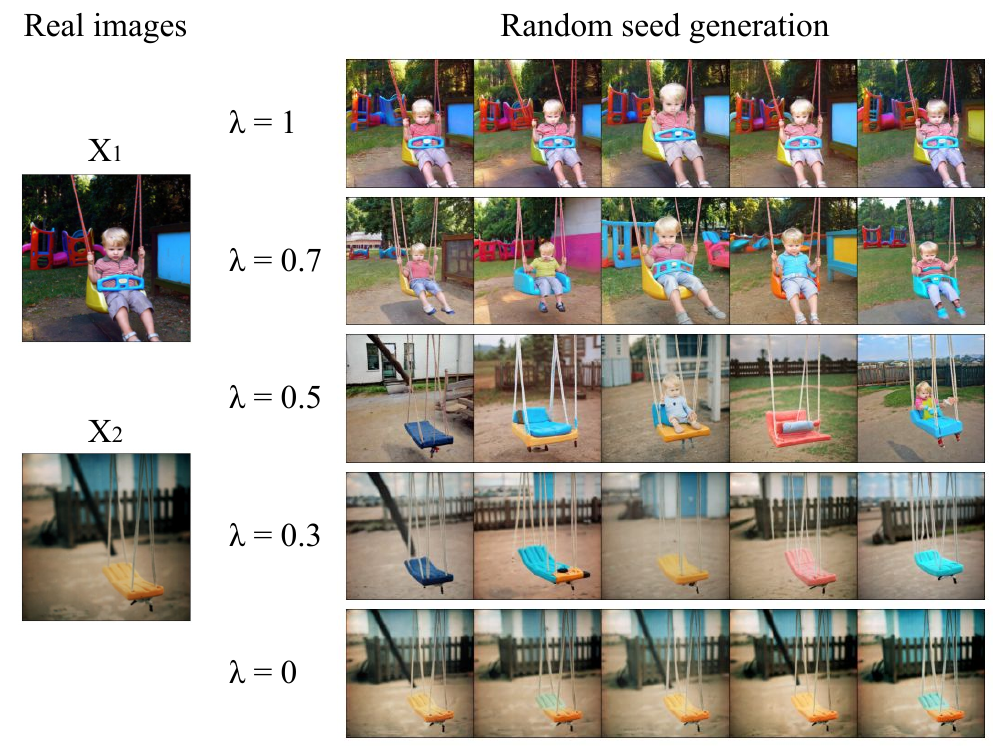}
      \caption{Swing}
      \label{fig:supp_lambda_sub4}
    \end{subfigure}
    \vspace{10pt}
    \caption{Ablation study of qualitative results on $\lambda$ variation when fusing LoRAs.} 
    \label{fig:supp_lambda_variation}
\end{figure*}

We present additional qualitative results in $\lambda$ variation for 4 classes, i.e. Hourglass, Carved Pumpkin, Diaper, and Swing, in Figure~\ref{fig:supp_lambda_variation}.

Taking the class Hourglass in Figure~\ref{fig:supp_lambda_sub1} as an example, a real image $x_1$ shows a single hourglass with a wooden frame while another real image $x_2$ shows multiple hourglasses without a wooden frame. When $\lambda=1$ or $0$, the images generated by different random seeds closely resemble one of the real images. When $\lambda=0.5$, the generated images show both diversity and high fidelity: some images have wooden frame while others do not, and some display multiple multiple hourglasses while others show only a single hourglass.

Taking the class Swing in Figure~\ref{fig:supp_lambda_sub4} as another example, a real image $x_1$ shows a baby riding a swing colored with yellow and blow while another real image $x_2$ shows only a yellow swing. When $\lambda=1$ or $0$, the images generated by different random seeds closely resemble one of the real images. When $\lambda=0.5$, the generated images show both diversity and high fidelity: the color of the swing is different, and a baby is riding a swing in some of the images.

\begin{figure*}
    \centering
    \includegraphics[width=\linewidth]{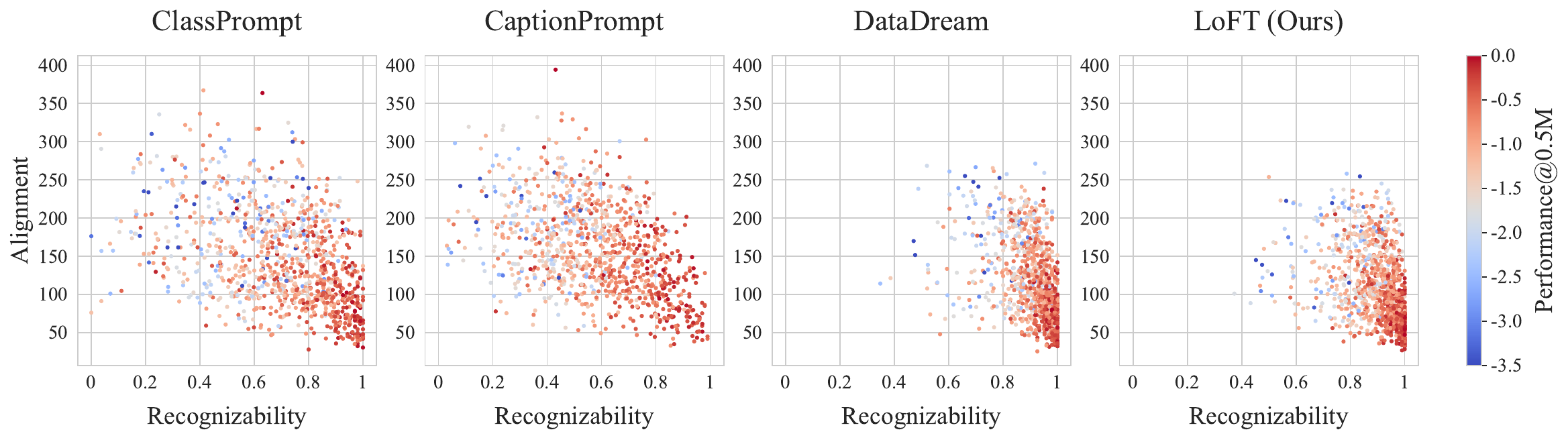}
    \vspace{-15pt}
    \caption{
    Per-class analysis of recognizability and alignment in synthetic datasets generated from different methods. The color indicates a log-likelihood of the ImageNet validation dataset when CLIP is fine-tuned on the 0.5M-sized synthetic dataset in the 16-shot setting.
    }
    \label{fig:supp_per_class_analysis_recog_fid}
\end{figure*}

\end{document}